%% file: main.tex
\pdfoutput=1

\documentclass[final]{cvpr}

\usepackage{times}
\usepackage{epsfig}
\usepackage{graphicx}
\usepackage{amsmath}
\usepackage{amssymb}
\usepackage{tabu}
\usepackage{booktabs}
\usepackage{multirow}
\usepackage[dvipsnames]{xcolor}

\usepackage[pagebackref=true,breaklinks=true,colorlinks,bookmarks=false]{hyperref}

\def\cvprPaperID{1212} 
\def\confYear{CVPR 2021}

\begin{document}

\title{Offboard 3D Object Detection from Point Cloud Sequences}


\author{%
Charles R. Qi~~~~Yin Zhou~~~~Mahyar Najibi~~~~Pei Sun~~~~Khoa Vo~~~~Boyang Deng~~~~Dragomir Anguelov\vspace{0.2cm}\\
Waymo LLC \\
}

\maketitle

\begin{abstract}

\input{./tex/abstract}
\end{abstract}




\section{Introduction}
\input{./tex/introduction}

\section{Related Work}
\input{./tex/related_work}

\section{Offboard 3D Object Detection}
\label{sec:problem}

\input{./tex/problem}

\section{3D Auto Labeling Pipeline} 
\label{sec:method}
\input{./tex/method}
\section{Experiments}
\input{./tex/experiment}

\section{Conclusion}
\input{./tex/conclusion}

{\small
\bibliographystyle{ieee_fullname}
\bibliography{pcl,yin_ref}
}

\newpage
\appendix
\section*{Appendix}
\input{./tex/supplementary}


\end{document}

%% file: tex/abstract.tex
While current 3D object recognition research mostly focuses on the real-time, onboard scenario, there are many offboard use cases of perception that are largely under-explored, such as using machines to automatically generate high-quality 3D labels. Existing 3D object detectors fail to satisfy the high-quality requirement for offboard uses due to the limited input and speed constraints. In this paper, we propose a novel offboard 3D object detection pipeline using point cloud sequence data. Observing that different frames capture complementary views of objects, we design the offboard detector to make use of the temporal points through both multi-frame object detection and novel object-centric refinement models.
Evaluated on the Waymo Open Dataset, our pipeline named 3D Auto Labeling shows significant gains compared to the state-of-the-art onboard detectors and our offboard baselines. Its performance is even on par with human labels verified through a human label study. Further experiments demonstrate the application of auto labels for semi-supervised learning and provide extensive analysis to validate various design choices.

%% file: tex/introduction.tex

\begin{figure}[t]
    \centering
    \includegraphics[width=0.9\linewidth]{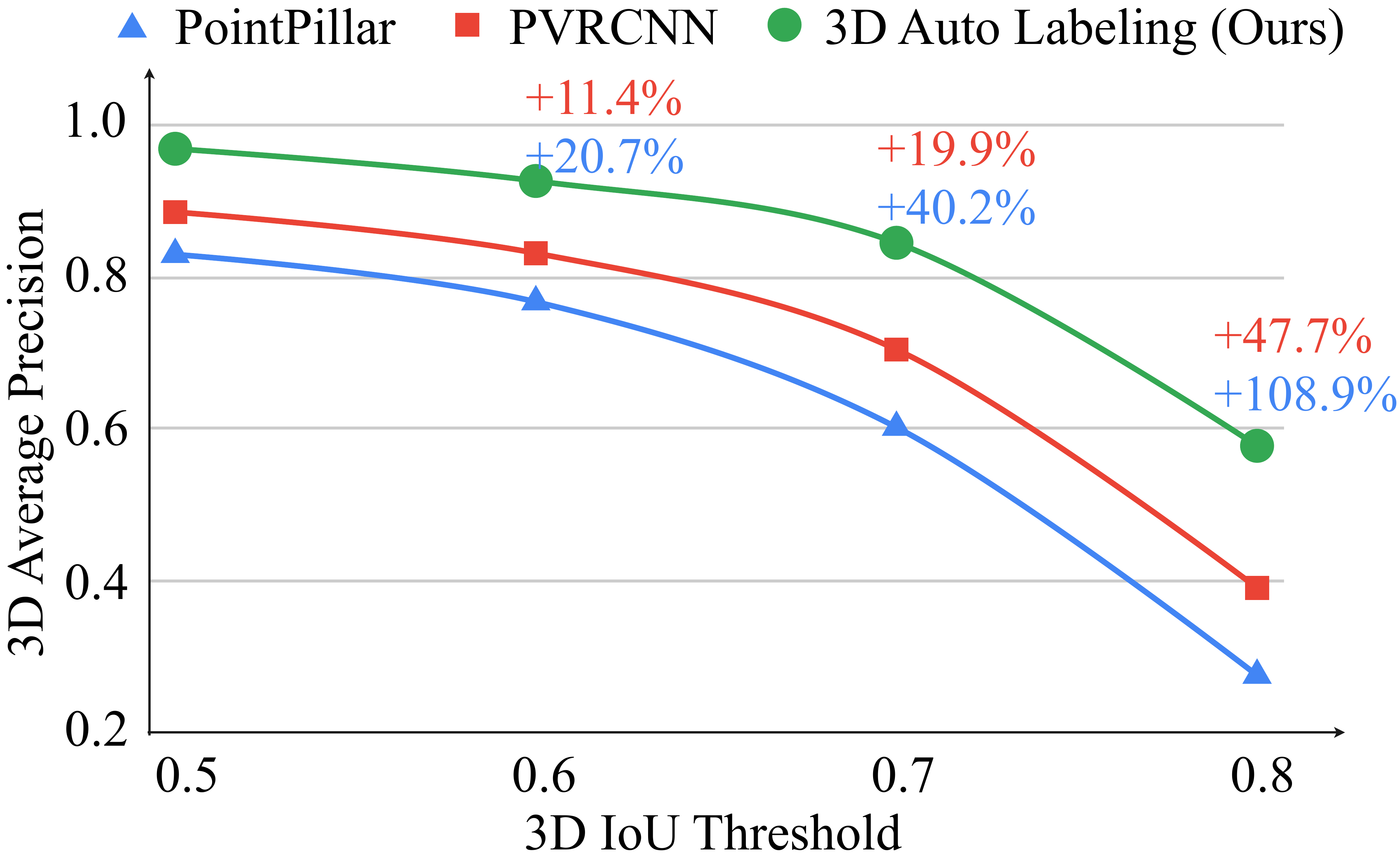}
    \caption{\textbf{Our offboard 3D Auto Labeling achieved significant gains over two representative onboard 3D detectors} (the efficient PointPillar~\cite{lang2019pointpillars} and the top-performing PVRCNN~\cite{shi2020pv}). The relative gains (the percentage numbers) are higher under more strict standard (higher IoU thresholds). The metric is 3D AP (L1) for vehicles on the Waymo Open Dataset~\cite{sun2020scalability} \emph{val} set.}
    \label{fig:teaser}
\end{figure}

Recent years have seen a rapid progress of 3D object recognition with advances in 3D deep learning and strong application demands. However, most 3D perception research has been focusing on real-time, onboard use cases and only considers sensor input from the current frame or a few history frames. Those models are sub-optimal for many \emph{offboard} use cases where the best perception quality is needed.
Among them, one important direction is to have machines ``auto label'' the data to save the cost of human labeling. High quality perception can also be used for simulation or to build datasets to supervise or evaluate downstream modules such as behavior prediction.


In this paper, we propose a novel pipeline for offboard 3D object detection with a modular design and a series of tailored deep network models. The offboard pipeline makes use of the whole sensor sequence input (such video data is common in applications of autonomous driving and augmented reality). With no constraints on the model causality and little constraint on model inference speed, we are able to greatly expand the design space of 3D object detectors and achieve significantly higher performance. 

\begin{figure}[t!]
    \centering
    \includegraphics[width=0.85\linewidth]{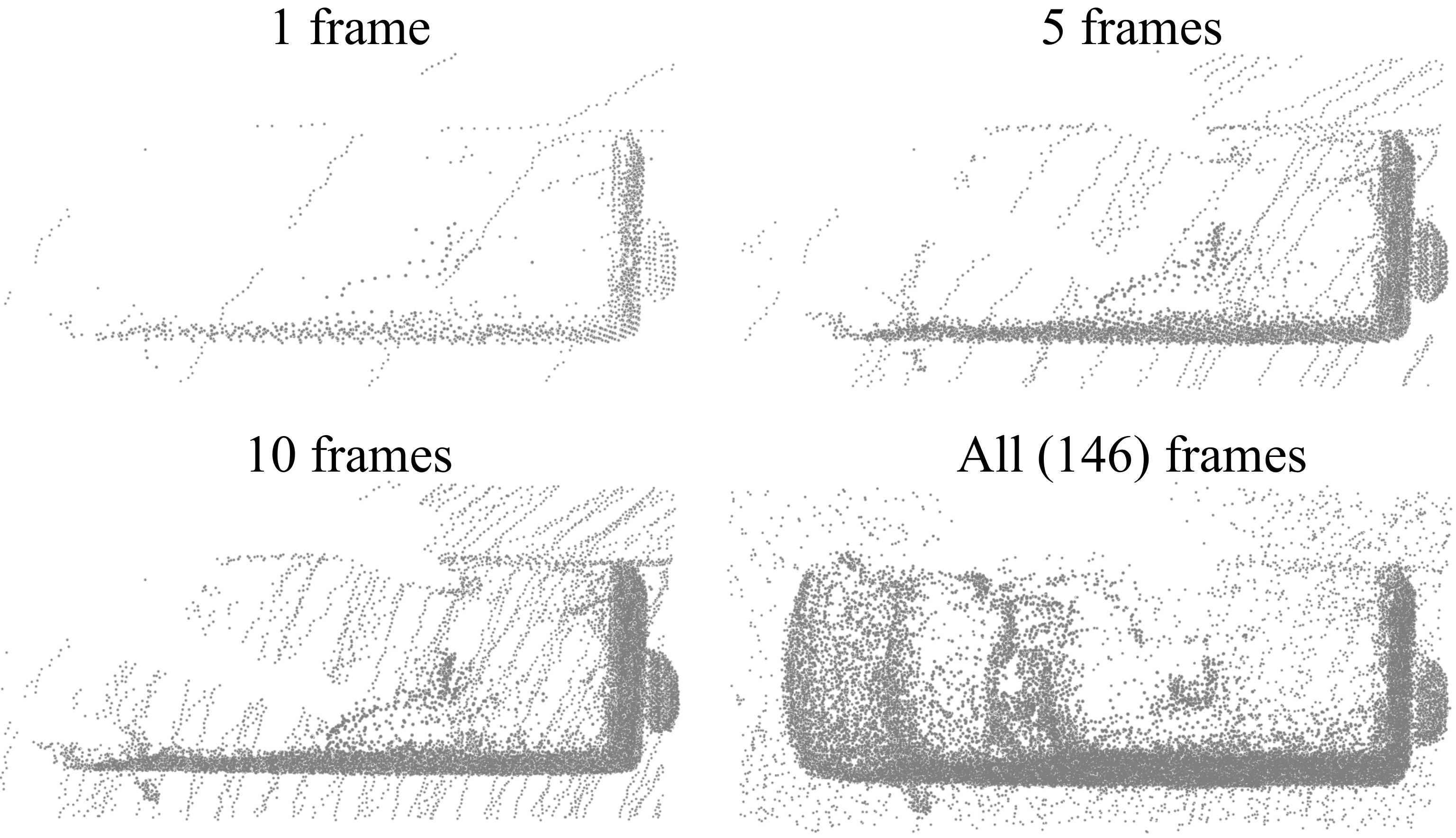}
    \caption{\textbf{Illustration of the complementary views of an object from the point cloud sequence.}
    Point clouds (aggregated from multiple frames) visualized in a top-down view for a mini-van.}
    \label{fig:point_aggregation}
\end{figure}

We design our offboard 3D detector based on a key observation: different viewpoints of an object, within a point cloud sequence, contain complementary information about its geometry (Fig.~\ref{fig:point_aggregation}). An immediate baseline design is to extend the current detectors to use multi-frame inputs. 
However, as multi-frame detectors are effective they are still limited in the amount of context they can use and are not naively scalable to more frames – gains from adding more frames diminish quickly (Table~\ref{tab:detection_AP_vs_frames}).

\begin{figure*}[t]
    \centering
    \includegraphics[width=0.95\linewidth]{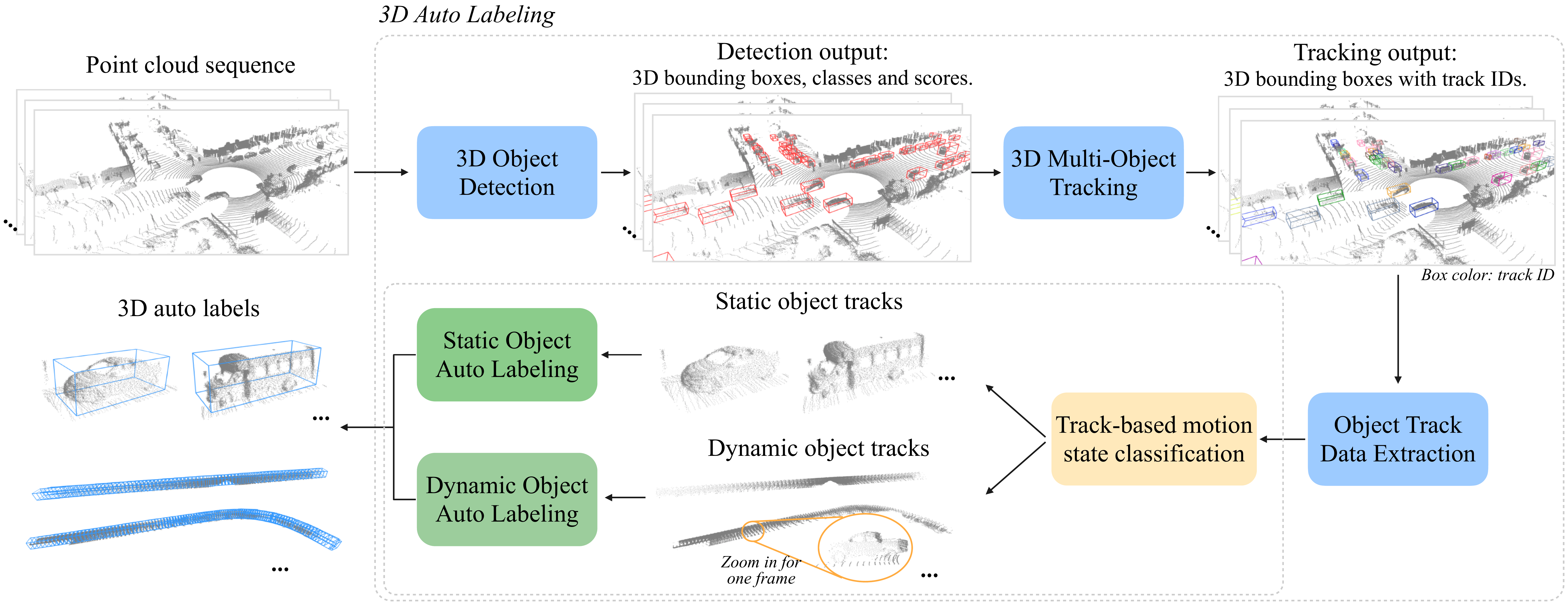}
    \caption{\textbf{The 3D Auto Labeling pipeline.} Given a point cloud sequence as input, the pipeline first leverages a 3D object detector to localize objects in each frame. Then object boxes at different frames are linked through a multi-object tracker. Object track data (its point clouds at every frame as well as its 3D bounding boxes) are extracted for each object and then go through the object-centric auto labeling (with a divide-and-conquer for static and dynamic tracks) to generate the final ``auto labels'', \ie refined 3D bounding boxes.}
    \label{fig:pipeline}
\end{figure*}

In order to fully utilize temporal point clouds (\eg 10 or more seconds), we step away from the common frame-based input structure where the entire frames of point clouds are merged. Instead, we turn to an \emph{object-centric} design. We first leverage a top-performing multi-frame detector to give us initial object localization. Then, we link objects detected at different frames through multi-object tracking. Based on the tracked boxes and the raw point cloud sequences, we can extract the entire track data of an object, including all of its sensor data (point clouds) and detector boxes, which is 4D: 3D spatial plus 1D temporal. We then propose novel deep network models to process such 4D object track data and output temporally consistent and high-quality boxes of the object. As they are similar to how a human labels an object and because of their high-quality output, we call those models processing the 4D track data as ``object-centric auto labeling models'' and the entire pipeline ``3D Auto Labeling'' (Fig.~\ref{fig:pipeline}).

We evaluate our proposed models on the Waymo Open Dataset (WOD)~\cite{sun2020scalability} which is a large-scale autonomous driving benchmark containing 1,000+ Lidar scan sequences with 3D annotations for every frame. Our 3D Auto Labeling pipeline dramatically lifts the perception quality compared to existing 3D detectors designed for the real-time, onboard use cases (Fig.~\ref{fig:teaser} and Sec.~\ref{sec:exp:sota}). The gains are even more significant at higher standards.
To understand how far we are from human performance in 3D object detection, we have conducted a human label study to compare auto labels with human labels (Sec.~\ref{sec:exp:human}). To our delight, we found that auto labels are already on par or even slightly better compared to human labels on the selected test segments.

In Sec.~\ref{sec:exp:semi_supervised}, we demonstrate the application of our pipeline for semi-supervised learning and show significantly improved student models trained with auto labels. We also conduct extensive ablation and analysis experiments to validate our design choices in Sec.~\ref{sec:exp:multi_frame_detector} and Sec.~\ref{sec:exp:analysis} and provide visualization results in Sec.~\ref{sec:exp:visualization}.

In summary, the contributions of our work are:
\begin{itemize}
    \item Formulation of the offboard 3D object detection problem and proposal of a specific pipeline (3D Auto Labeling) that leverages our multi-frame detector and novel object-centric auto labeling models.
    \item State-of-the-art 3D object detection performance on the challenging Waymo Open Dataset.
    \item The human label study on 3D object detection with comparisons between human and auto labels.
    \item Demonstrated the effectiveness of auto labels for semi-supervised learning.
\end{itemize}

%% file: tex/related_work.tex



\paragraph{3D object detection} Most work has been focusing on using single-frame input. In terms of the representations used, they can be categorized into voxel-based~\cite{REF:VotingforVoting_RSS2015,REF:Vote3Deep_ICRA2017,REF:3DFCN_RSJ2017,song2016deep,REF:ku2018joint,REF:yang2018pixor,simony2018complex,zhou2018voxelnet,REF:second_2018,lang2019pointpillars,REF:HVNet2020,REF:PillarNet_ECCV2020}, point-based~\cite{shi2018pointrcnn,REF:yang2018ipod,REF:StarNet_2019,qi2019deep,yang20203dssd,REF:Point-GNN_CVPR2020}, perspective-view-based~\cite{REF:VeloFCN2016,REF:lasernet_CVPR2019,REF:bewley2020range} as well as hybrid strategy~\cite{zhou2020end,REF:STD_ICCV2019,REF:FastPointRCNN_Jiaya_ICCV2019,REF:SA_SSD_He_2020_CVPR,shi2020pv}.
Several recent works explored temporal aggregation of Lidar scans for point cloud densification and shape completion. \cite{REF:FaF_Luo_2018_CVPR} fuses multi-frame information by concatenating feature maps from different frames. \cite{REF:you_see_Hu_2020_CVPR} aggregates (motion-compensated) points from different Lidar sweeps into a single scene. \cite{REF:Spatiotemporal_transformer_CVPR2020} uses graph-based spatiotemporal feature encoding to enable message passing among different frames. \cite{REF:ConvLSTM_ECCV2020} encodes previous frames with a LSTM to assist detection in the current frame.
Using multi-modal input (camera views and 3D point clouds)~\cite{REF:ku2018joint,cvpr17chen,qi2018frustum,REF:pointfusion_CVPR2018,REF:ContFuse_ECCV2018,REF:Multi_task_multisensor_fusion_CVPR2019,REF:LaserNet++_CVPRW2019,REF:MVX-Net_ICRA2019,qi2020imvotenet} has shown improved 3D detection performance compared to point-cloud-only methods, especially for small and far-away objects.
In this work, we focus on a point-cloud-only solution and on leveraging data over a long temporal interval. 

\paragraph{Learning from point cloud sequences} Several recent works~\cite{liu2019flownet3d, gu2019hplflownet, mittal2020just} proposed to learn to estimate scene flow from dynamic point clouds using end-to-end trained deep neural networks (from a pair of consecutive point clouds). Extending such ideas, MeteorNet~\cite{liu2019meteornet} showed that longer sequences input can lead to performance gains for tasks such as action recognition, semantic segmentation and scene flow estimation. There are also other applications of learning in point cloud sequences, like point cloud completion~\cite{prantl2019tranquil}, future point cloud prediction~\cite{weng20204d} and gesture recognition~\cite{owoyemi2018spatiotemporal}. We also see more released datasets with sequence point cloud data such as the Waymo Open Dataset~\cite{sun2020scalability} for detection and the SemanticKITTI dataset~\cite{behley2019semantickitti} for 3D semantic segmentation.

\paragraph{Auto labeling}

The large datasets required for training data-hungry models have increased the annotation costs noticeably in recent years. Accurate auto labeling can dramatically reduce annotation time and cost. 
Previous works on auto labeling were mainly focused on 2D applications. Lee \etal proposed pseudo-labeling~\cite{lee2013pseudo} to use the most confident predicted category of an image classifier as labels to train it on the unlabeled part of the dataset. More recent works~\cite{iscen2019label, zou2019confidence, yalniz2019billion, xie2020self} have further improved the procedures to use pseudo labels and demonstrated wide success including state-of-the-art results on ImageNet~\cite{deng2009imagenet}.

For 3D object detection, recently, Zakharov \etal~\cite{zakharov2020autolabeling} proposed an auto labeling framework using pre-trained 2D detectors to annotate 3D objects. While effective for loose localization (\ie IoU of 0.5), there is a considerable performance gap for applications requiring higher precision. \cite{meng2020weakly} tried to leverage weak center-click supervision to reduce 3D labels needed.
Several other works~\cite{castrejon2017annotating, acuna2018efficient, lee2018leveraging, ling2019fast, feng2019deep} have also proposed methods to assist human annotators and consequently reducing the annotation cost.







%% file: tex/problem.tex
\paragraph{Problem statement}
Given a sequence of sensor inputs (temporal data) of a dynamic environment, our goal is to localize and classify objects in the 3D scene for every frame.
Specifically, we consider the input of a sequence of point clouds $\{\mathcal{P}_i \in \mathbf{R}^{n_i \times C}\}$, $i=1,2,...,N$ with the point cloud $\mathcal{P}_i$ ($n_i$ points with $C$ channels for each point) of each of the $N$ total frames.
The point channels include the $XYZ$ in the sensor's coordinate (at each frame) and other optional information such as color and intensity.
We also assume known sensor poses $\{\mathcal{M}_i = [R_i | t_i] \in \mathbf{R}^{3\times 4}\}$, $i=1,2,...,N$ at each frame in the world coordinate, such that we can compensate the ego-motion.
For each frame, we output amodal 3D bounding boxes (parameterized by its center, size and orientation), class types (\eg vehicles) and unique object IDs for all objects that appear in the frame.

\paragraph{Design space} Access to temporal data (history and future) has led to a much larger design space of detectors compared to just using single frame input. 



One baseline design is to extend the single-frame 3D object detectors to use multi-frame input. Although previous works~\cite{REF:FaF_Luo_2018_CVPR, REF:you_see_Hu_2020_CVPR, REF:ConvLSTM_ECCV2020, REF:Spatiotemporal_transformer_CVPR2020} have shown its effectiveness, a multi-frame detector is hard to scale up to more than a few frames and cannot compensate the object motions since frame stacking is done for the entire scene. We observe that the contributions of multi-frame input to the detector quality diminish as we stack more frames (Table~\ref{tab:detection_AP_vs_frames}).
Another idea is to extend the second stage of two-stage detectors~\cite{qi2018frustum, shi2018pointrcnn} to take object points from multiple frames. Compared to taking multi-frame input of the whole scene, the second-stage only processes proposed object regions. However, it is not intuitive to decide how many context frames to use. Setting a fixed number may work well for some objects but suboptimal for others.


Compared to the \emph{frame-centric} designs above, where input is always from a fixed number of frames, we recognize the necessity to adaptively choose the temporal context size for each object independently, leading to an \emph{object-centric} design. As shown in Fig.~\ref{fig:pipeline}, we can leverage the powerful multi-frame detector to give us the initial object localizations. Then for each object, through tracking, we can extract all relevant object point clouds and detection boxes from all frames that it appears in. Subsequent models can take such object track data to output the final track-level refined boxes of the objects. As this process emulates how a human labeler annotates a 3D object in the point cloud sequence (localize, track and refine the track over time), we chose to refer to our pipeline as \emph{3D Auto Labeling}.


%% file: tex/method.tex



Fig.~\ref{fig:pipeline} illustrates our proposed 3D Auto Labeling pipeline. We will introduce each module of the pipeline in the following sub-sections.

\subsection{Multi-frame 3D Object Detection}
\paragraph{MVF++}
As the entry point to our pipeline, accurate object detection is essential for the downstream modules. 
In this work, we propose the MVF++ 3D detector by extending the top-performing Multi-View Fusion~\cite{zhou2020end} (MVF) detector in three aspects: 1) to enhance the discriminative ability of point-level features, we add an auxiliary loss for 3D semantic segmentation, where points are labeled as positives/negatives if they lie inside/outside of a ground truth 3D box; 2) for obtaining more accurate training targets and improving training efficiency, we eliminate the anchor matching step in the MVF paper and adopt the anchor-free design as in~\cite{REF:tian2019fcos}; 3) to leverage ample computational resources available in the offboard setting, we redesign the network architecture and increase the model capacity. Please see Sec.~\ref{sec:supp:mvf} in the Appendix for details.

\paragraph{Multi-frame MVF++}
We extend the MVF model to use multiple LiDAR scans. Points from multiple consecutive scans are transformed to the current frame based on ego-motion. Each point is extended by one additional channel, encoding of the relative temporal offset, similar to~\cite{REF:you_see_Hu_2020_CVPR}. The aggregated point cloud is used as the input to the MVF++.

\paragraph{Test-time augmentation}
\label{sec:test_time_augmentation}
We further boost the 3D detection through test-time augmentation (TTA)~\cite{REF:AlexNet:2017}, by rotating the point cloud around Z-axis by 10 different angles (\ie [0, $\pm 1/8\pi$, $\pm 1/4\pi$, $\pm 3/4\pi$, $\pm 7/8\pi$, $\pi$]), and ensembling predictions with weighted box fusion~\cite{REF:wbf2019}. While it may lead to excessive computational complexity for onboard uses, in the offboard setting TTA can be parallelized across multiple devices for fast execution. 



\subsection{Multi-object Tracking}

The multi-object tracking module links detected objects across frames.
Given the powerful multi-frame detector, we choose to take the tracking-by-detection path and have a separate non-parametric tracker. This leads to a simpler and more modular design compared to the joint detection and tracking methods~\cite{REF:FaF_Luo_2018_CVPR,weng2020joint,li2020joint}.
Our tracker is an implementation variant of the~\cite{weng2019baseline}, using detector boxes for associations and Kalman filter for state updates.

\subsection{Object Track Data Extraction}
Given tracked detection boxes for an object, we can extract object-specific LiDAR point clouds from the sequence. We use the term \emph{object track data} to refer to such 4D (3D spatial and 1D temporal) object information.

To extract object track data, we first transform all boxes and point clouds to the world coordinate through the known sensor poses to remove the ego-motion. For each unique object (according to the object ID), we crop its object points within the estimated detector boxes (enlarged by $\alpha$ meters in each direction to include more contexts). 
Such extraction gives us a sequence of object point clouds $\{\mathcal{P}_{j, k}\}$, $k \in S_j$ for each object $j$ and its visible frames $S_j$. Fig.~\ref{fig:pipeline} visualizes the object points for several vehicles. Besides the raw point clouds, we also extract the tracked boxes for each object and every frame $\{\mathcal{B}_{j,k}\}$, $k \in S_j$ in the world coordinate.

\begin{figure}[t!]
    \centering
    \includegraphics[width=\linewidth]{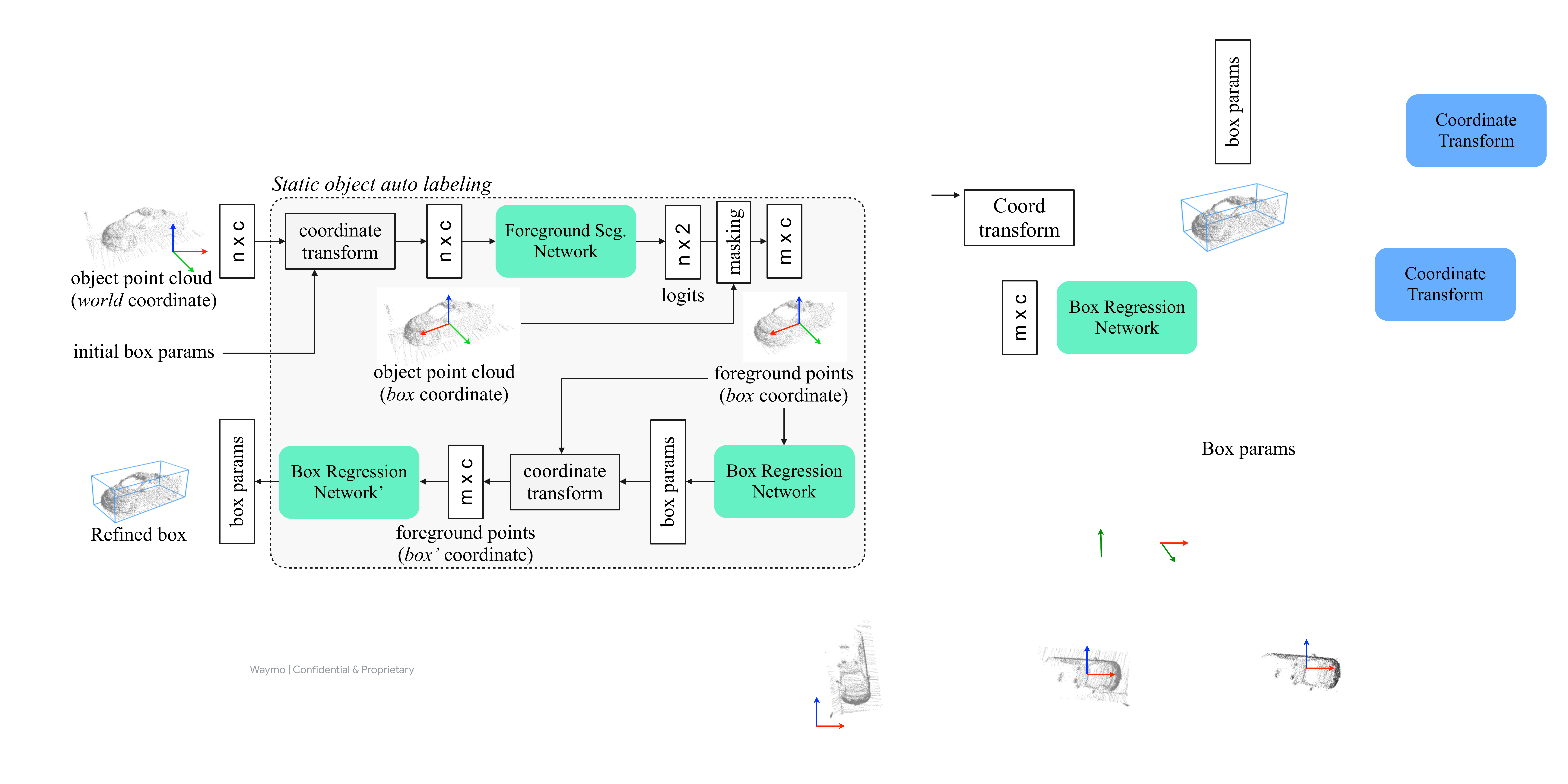}
    \caption{\textbf{The static object auto labeling model.} Taking as input the merged object points in the world coordinate, the model outputs a single box for the static object.}
    \label{fig:static_model}
\end{figure}

\subsection{Object-centric Auto Labeling}
In this section, we describe how we take the object track data to ``auto label'' the objects. As illustrated in Fig.~\ref{fig:pipeline}, the process includes three sub-modules: the track-based motion state classification, static object auto labeling and dynamic object auto labeling, which are described in detail below.

\paragraph{Divide and conquer: motion state estimation}


In the real world, lots of objects are completely static during a period of time. For example, parked cars or furniture in a room do not move within a few minutes or hours. In terms of offboard detection, it is preferred to assign a single 3D bounding box to a static object rather than separate boxes in different frames to avoid jittering.

Based on this observation, we take a divide-and-conquer approach to handle static and moving objects differently, introducing a module to classify an object's motion state (static or not) before the auto labeling.
While it could be hard to predict an object's motion state from just a few frames (due to the perception noise), we find it relatively easy if all object track data is used. As the visualization in Fig.~\ref{fig:pipeline} shows, it is often obvious to tell whether an object is static or not from its trajectory. A linear classifier using a few heuristic features from the object track's boxes can already achieve $99\%+$ motion state classification accuracy for vehicles. More details are in Sec.~\ref{sec:supp:motion}.


\paragraph{Static object auto labeling}


For a static object, the model takes the merged object point clouds ($\mathcal{P}_j = \cup \{\mathcal{P}_{j, k}\}$ in the world coordinate) from points at different frames and predicts a single box. The box can then be transformed to each frame through the known sensor poses.

Fig.~\ref{fig:static_model} illustrates our proposed model for static object auto labeling.
Similar to~\cite{qi2018frustum, shi2018pointrcnn}, we first transform (through rotation and translation) the object points to a \emph{box coordinate} before the per-object processing, such that the point clouds are more aligned across objects. In the box coordinate, the $+X$ axis is the box heading direction, the origin is the box center.
Since we have the complete sequence of the detector boxes, we have multiple options on which box to use as the initial box. The choice actually has a significant impact on model performance. Empirically, using the box with the highest detector score leads to the best performance (see Sec.~\ref{sec:supp:analysis} for an ablation study).

To attend to the object, the object points are passed through an instance segmentation network to segment the foreground ($m$ foreground points are extracted by the mask). Inspired by the Cascade-RCNN~\cite{cai2018cascade}, we iteratively regress the object's bounding box.
At test time, we can further improve box regression accuracy by test-time-augmentation (similar to Sec. \ref{sec:test_time_augmentation}).

All networks are based on the PointNet~\cite{qi2017pointnet} architecture. The model is supervised by the segmentation and box estimation ground truths. Details of the architecture, losses and the training process are described in Sec.~\ref{sec:supp:object_auto_labeling}.

\begin{figure}[t!]
    \centering
    \includegraphics[width=\linewidth]{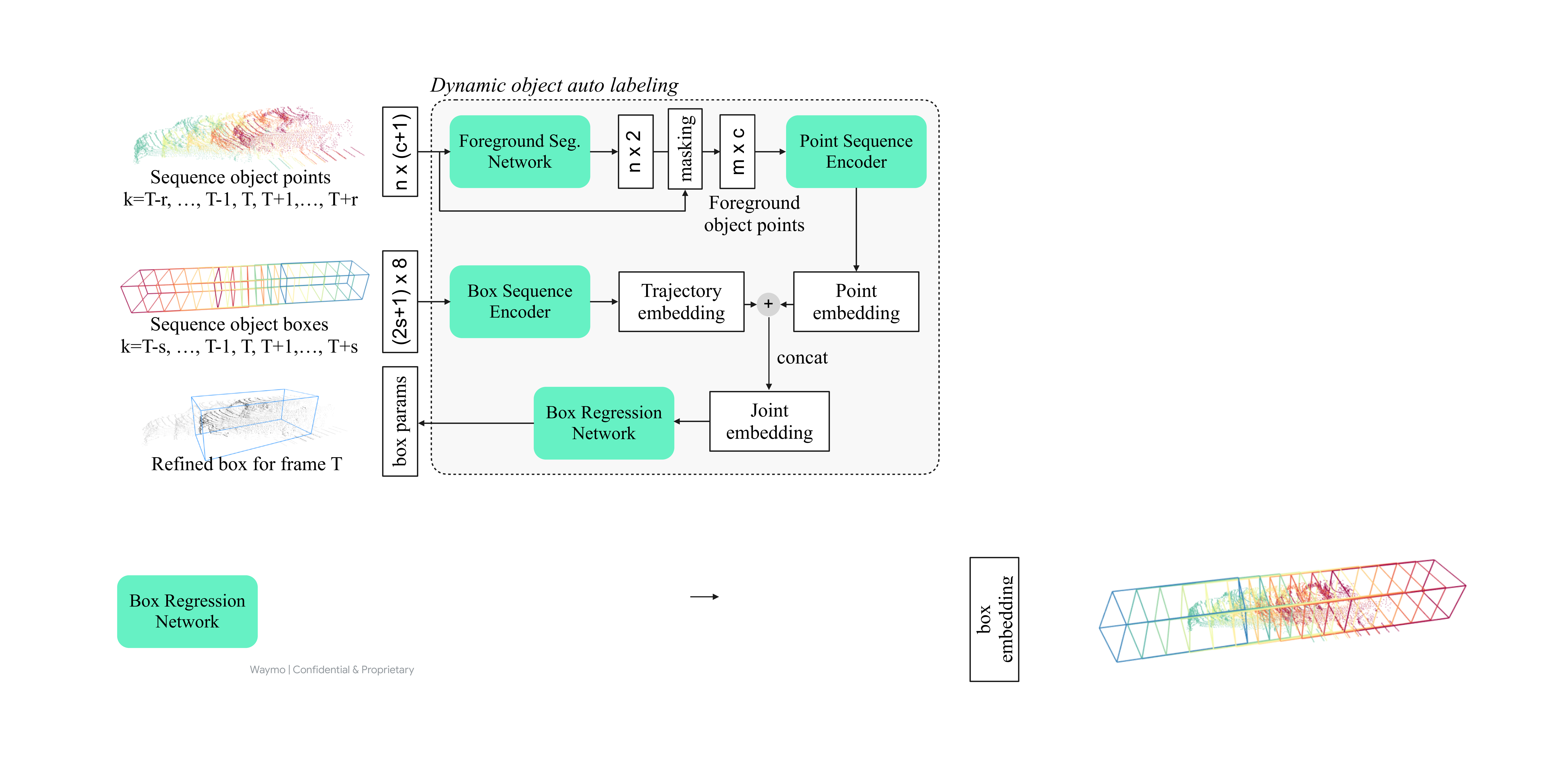}
    \caption{\textbf{The dynamic object auto labeling model.} Taking a sequence of object points and a sequence of object boxes, the model runs in a sliding window fashion and outputs a refined 3D box for the center frame. Input point and box colors represent frames.}
    \label{fig:dynamic_model}
\end{figure}

\begin{table*}[t!]
    \centering
    \setlength{\tabcolsep}{3pt}
    \small
    \begin{tabular}{l|c|c|c|c|c||c|c|c|c}
    \toprule
         \multirow{3}{*}{Method} & \multirow{3}{*}{frames} & \multicolumn{4}{c||}{\emph{Vehicles}} & \multicolumn{4}{c}{\emph{Pedestrians}} \\ 
         & & \multicolumn{2}{c|}{3D AP} & \multicolumn{2}{c||}{BEV AP} & \multicolumn{2}{c|}{3D AP} & \multicolumn{2}{c}{BEV AP} \\  
         & & IoU=0.7 & IoU=0.8 & IoU=0.7 & IoU=0.8 & IoU=0.5 & IoU=0.6 & IoU=0.5 & IoU=0.6\\ 
         \toprule
         StarNet~\cite{REF:StarNet_2019} & 1 & 53.70 & - & - & - & 66.80 & - & - & -\\
         PointPillar~\cite{lang2019pointpillars}$^{\star}$ & 1 & 60.25 & 27.67 & 78.14 & 63.79 &60.11 & 40.35 & 65.42 & 51.71 \\
         Multi-view fusion (MVF)~\cite{zhou2020end} & 1 & 62.93 & - & 80.40 & - & 65.33 & - & 74.38 & -\\
         AFDET~\cite{REF:AFDET_CVPRW2020} & 1 & 63.69 & - & - & - & - & - & - & - \\
         ConvLSTM~\cite{REF:ConvLSTM_ECCV2020} & 4 & 63.60  & - & - & - & - & - & - & -\\
         RCD~\cite{REF:bewley2020range} & 1 & 68.95 & - & 82.09 & - & - & - & - & -\\
         PillarNet~\cite{REF:PillarNet_ECCV2020} & 1 & 69.80 & - & 87.11 & - & 72.51 & - & 78.53 & -\\
         PV-RCNN~\cite{shi2020pv}$^{\star}$ & 1 & 70.47  & 39.16 & 83.43  & 69.52 & 65.34 & 45.12 & 70.35 & 56.63\\
         \midrule
         Single-frame MVF++ (Ours) & 1 & 74.64  & 43.30  & 87.59  & 75.30 & 78.01 & 56.02 & 83.31  & 68.04 \\
         Multi-frame MVF++ w. TTA (Ours) & 5 & 79.73  & 49.43  & 91.93  & 80.33 & 81.83 & 60.56 & 85.90 & 73.00 \\
        3D Auto Labeling (Ours) & all & \textbf{84.50} & \textbf{57.82}  & \textbf{93.30}  & \textbf{84.88} & \textbf{82.88} & \textbf{63.69} & \textbf{86.32} & \textbf{75.60} \\ 
         \bottomrule
    \end{tabular}
    \caption{\textbf{3D object detection results for vehicles and pedestrians on the Waymo Open Dataset \emph{val} set.} Methods in comparison include prior state-of-the-art single-frame based 3D detectors as well as our single-frame MVF++, our multi-frame MVF++ (5 frames) and our full 3D Auto Labeling pipeline. The metrics are L1 3D AP and bird's eye view (BEV) AP at two IoU thresholds: the common standard IoU=0.7 and a high standard IoU=0.8 for vehicles; and IoU = 0.5, 0.6 for pedestrians.
    $^{\star}$ reproduced results using author's released code.}
    \label{tab:main_detection_results}
\end{table*}

\paragraph{Dynamic object auto labeling}

For a moving object, we need to predict different 3D bounding boxes for each frame.
Due to the sequence input/output, the model design space is much larger than that for static objects.
A baseline is to re-estimate the 3D bounding box with cropped point clouds.
Similar to the smoothing in tracking, we can also refine boxes based on the sequence of the detector boxes.
Another choice is to ``align'' or register object points with respect to a keyframe (\eg the current frame) to obtain a denser point cloud for box estimation.
However, the alignment can be a harder problem than box estimation especially for occluded or faraway objects with fewer points. Besides, it is challenging to align deformable objects like pedestrians.

We propose a design (Fig.~\ref{fig:dynamic_model}) that leverages both the point cloud and the detector box sequences without aligning points to a keyframe explicitly.
Given a sequence of object point clouds $\{\mathcal{P}_{j, k}\}$ and a sequence of detector boxes $\{\mathcal{B}_{j, k}\}$ for the object $j$ at frames $k \in S_j$, the model predicts the object box at each frame $k$ in a sliding window form. It consists of two branches, one taking the point sequence and the other taking the box sequence.


For the point cloud branch, the model takes a sub-sequence of the object point clouds $\{\mathcal{P}_{j,k}\}_{k=T-r}^{T+r}$. After adding a temporal encoding channel to each point (similar to~\cite{REF:you_see_Hu_2020_CVPR})
, the sub-sequence points are merged through union and transformed to the box coordinate of the detector box $\mathcal{B}_{j,T}$ at the center frame. Following that, we have a PointNet~\cite{qi2017pointnet} based segmentation network to classify the foreground points (of the $2r+1$ frames) and then encode the object points into an embedding through another point encoder network.

For the box sequence branch, the box sequences $\{\mathcal{B}'_{j,k}\}_{k=T-s}^{T+s}$ of $2s+1$ frames are transformed to the box coordinate of the detector box at frame $T$. Note that the box sub-sequence can be longer than the point sub-sequence to capture the longer trajectory shape. A box sequence encoder network (a PointNet variant) will then encode the box sequence into a trajectory embedding, where each box is a point with 7-dim geometry and 1-dim time encoding.

Next, the computed object embedding and the trajectory embedding are concatenated to form the joint embedding which will then be passed through a box regression network to predict the object box at frame $T$.

%% file: tex/experiment.tex

We start the section by comparing our offboard 3D Auto Labeling with state-of-the-art 3D object detectors in Sec.~\ref{sec:exp:sota}. In Sec.~\ref{sec:exp:human} we compare the auto labels with the human labels. In Sec.~\ref{sec:exp:semi_supervised}, we show how the auto labels can be used to supervise a student model to achieve improved performance under low-label regime or in another domain. We provide analysis of the multi-frame detector in Sec.~\ref{sec:exp:multi_frame_detector} and analysis experiments to validate our designs of the object-centric auto labeling models in Sec.~\ref{sec:exp:analysis} and finally visualize the results in Sec.~\ref{sec:exp:visualization}.

\paragraph{Dataset}
\label{sec:exp:dataset}
We evaluate our approach using the challenging Waymo Open Dataset (WOD)~\cite{sun2020scalability}, as it provides a large collection of LiDAR sequences, with 3D labels available for each frame. The dataset includes a total number of 1150 sequences with 798 for training, 202 for validation and 150 for testing. Each LiDAR sequence lasts around 20 seconds with a sampling frequency at 10Hz.
For our experiments, we evaluate both 3D and bird's eye view (BEV) object detection metrics for vehicles and pedestrians.



\subsection{Comparing with State-of-the-art Detectors}
\label{sec:exp:sota}



In Table~\ref{tab:main_detection_results}, we show comparisons of our 3D object detectors and the 3D Auto Labeling with various single-frame and multi-frame based detectors, under both the common standard IoU threshold and a higher standard IoU threshold to pressure test the models.


We show that our single-frame MVF++ has already outperformed the prior art single-frame detector PVRCNN~\cite{shi2020pv}. The multi-frame version of the MVF++, as a baseline of the offboard 3D detection methods, significantly improves upon the single-frame MVF++ thanks to the extra information from the context frames.

For vehicles, comparing the last three rows, our complete 3D Auto Labeling pipeline, which leverages the multi-frame MVF++ and the object-centric auto labeling models, further improves the detection quality especially in the higher standard at IoU threshold of 0.8. It improves the 3D AP@0.8 significantly by $\textbf{14.52}$ points compared to the single-frame MVF++ and by $\textbf{8.39}$ points compared to the multi-frame MVF++, which is already very powerful by itself. These results show the great potential of leveraging the long sequences of point clouds for offboard perception.

We also show the detection AP for the pedestrian class, where we consistently observe the leading performance of the 3D Auto Labeling pipeline especially at the higher localization standard (IoU=0.6) with $\textbf{7.67}$ points gain compared to the single-frame MVF++ and $\textbf{3.13}$ points gain compared to the multi-frame MVF++.

\begin{table}[t!]
    \centering
    \small
    \begin{tabular}{l|cc|cc}
    \toprule
    & \multicolumn{2}{c|}{3D AP} & \multicolumn{2}{c}{BEV AP} \\
          & IoU=0.7 & IoU=0.8 & IoU=0.7 & IoU=0.8 \\ \toprule
         Human & \textbf{86.45} & \textbf{60.49} & \textbf{93.86} & 86.27 \\ 
         3DAL (Ours) & 85.37 & 56.93 & 92.80 & \textbf{87.55} \\ 
         \bottomrule
    \end{tabular}
    \caption{\textbf{Comparing human labels and auto labels in 3D object detection.} The metrics are 3D and BEV APs for vehicles on the 5 sequences from the Waymo Open Dataset \emph{val} set. Human APs are computed by comparing them with the WOD's released ground truth and using number of points in boxes as human label scores.}
    \label{tab:human_label_study}
\end{table}

\subsection{Comparing with Human Labels}
\label{sec:exp:human}

In many perception domains such as image classification and speech recognition, researchers have collected data to understand humans' capability~\cite{russakovsky2015imagenet,deshmukh1996benchmarking,lippmann1997speech}. However, to the best of our knowledge, no such study exists for 3D recognition especially for 3D object detection.
To fill this gap, we conducted a small-scale human label study on the Waymo Open Dataset to understand the capability of human in recognizing objects in a dynamic 3D scene. We randomly selected 5 sequences from the Waymo Open Dataset \emph{val} set and asked three experienced labelers to re-label each sequence independently (with the same labeling protocol as WOD).
In Table~\ref{tab:human_label_study}, we report the mean AP of human labels and auto labels across the 5 sequences.
With the common 3D AP@0.7 (L1) metric, the auto labels are only around 1 point lower than the average labeler, although the gap is slightly larger in the more strict 3D AP@0.8 metric. With some visualization, we found the larger gap is mostly caused by inaccurate heights. The comparisons with the BEV AP@0.8 metric verifies our observation: when we don't consider height, the auto labels even outperform the average human labels by $1.28$ points.

With such high quality, we believe the auto labels can be used to pre-label point cloud sequences to assist and accelerate human labeling, or be used directly to train light-weight student models as shown in the following section.

\begin{table}[t!]
    \centering
    \resizebox{\columnwidth}{!}{
    \begin{tabu}{c|c|c|c}
    \toprule
        Training Data & Test Data & 3D AP & BEV AP \\
        \toprule
        100\% main \textit{train} (Human) & main \textit{val} & 71.2  & 86.9  \\
        \tabucline [0.1pt on 4pt off 2pt]{1-4} 
        10\% main \textit{train} (Human) &  main \textit{val} & 64.3 & 81.2 \\
        \tabucline [0.1pt on 4pt off 2pt]{1-4} 
        
         10\% main \textit{train} (Human) & \multirow{2}{*}{main \textit{val}} & \multirow{2}{*}{70.0} & \multirow{2}{*}{86.4} \\
         + 90\% main \textit{train} (\textbf{3DAL}) & & & \\
         
         \midrule 
         
         100 \% main \textit{train} (Human) & domain \textit{test} & 59.4 & \textit{N/A} \\ 
         \tabucline [0.1pt on 4pt off 2pt]{1-4}
         100 \% main \textit{train} (Human) & \multirow{2}{*}{domain \textit{test}} & \multirow{2}{*}{60.3} & \multirow{2}{*}{\textit{N/A}} \\
         + domain  (Self Anno.) & & & \\
         \tabucline [0.1pt on 4pt off 2pt]{1-4}
         100 \% \textit{train} (Human) & \multirow{2}{*}{domain \textit{test}} & \multirow{2}{*}{64.2} & \multirow{2}{*}{\textit{N/A}} \\
         + domain  (\textbf{3DAL}) & & & \\
         \bottomrule
    \end{tabu}
    }
    \caption{\textbf{Results of semi-supervised learning with auto labels.} Metrics are 3D and BEV AP for vehicles on the Waymo Open Dataset. The type of annotation is reported in parenthesis. Please note, test set BEV AP is not provided by the submission server.}
    \label{tab:semi_supervisedv2}
\end{table}

\subsection{Applications to Semi-supervised Learning}
\label{sec:exp:semi_supervised}
In this section, we study the effectiveness of our auto labeling pipeline in the task of semi-supervised learning to train a student model under two settings: intra-domain and cross-domain. We choose the student model as a single-frame MVF++ detector that can run in real-time.

For the \emph{intra-domain semi-supervised learning},
we randomly select 10\% sequences (79 ones) in the main WOD training set to train our 3D Auto Labeling (3DAL) pipeline. Once trained, we apply it to the rest 90\% sequences (719 ones) in the main training set to generate ``auto labels'' (we only keep boxes with scores higher than $0.1$).
In Table~\ref{tab:semi_supervisedv2} (first two rows), we see that reducing the human annotations to 10\% significantly lowers the student model's performance. However, when we use auto labels, the student model trained on 10\% human labels and 90\% auto labels can get similar performance compared to using 100\% human labels (AP gaps smaller than 1 point), demonstrating superb data efficiency auto labels can provide.

For the \emph{cross-domain semi-supervised learning},
the teacher auto labels data from an unseen domain. The teacher is trained on the main WOD \emph{train} set, and auto labels the domain adaptation WOD \emph{train} and \emph{unlabeled} sets (separate 680 sequences from the main WOD). The student is then trained on the union of these three sets. Evaluations are on the domain adaptation \emph{test} set.
The last three rows of Table~\ref{tab:semi_supervisedv2} show the results.
Without using any data from the new domain, the student gets an AP of $59.4$.
While using the student to self-label slightly helps (improves the results by $\sim 1$ point), using our 3DAL to auto label the new domain significantly improves the student AP by $\sim 5$ points.

\subsection{Analysis of the Multi-frame Detector}
\label{sec:exp:multi_frame_detector}

Table~\ref{tab:detector_ablation} shows the ablations of our proposed MVF++ detectors. We see that the offboard techniques such as the model capacity increase ($+3.08$ AP@0.7), using point clouds from 5 frames as input ($+1.70$ AP@0.7) and test time augmentation ($+3.39$ AP@0.7) are all very effective in improving the detection quality.

Table~\ref{tab:detection_AP_vs_frames} shows how the number of consecutive input frames impacts the detection APs. The gains of adding frames quickly diminishes as the number of frames increases: \eg while the AP@0.8 improves by $0.81$ from 1 to 2 frames, the gain from 4 to 5 frames is only $0.14$ point.




\begin{table}[]
    \centering
    \scriptsize
    \begin{tabular}{c|c|c|c|c|c}
    \toprule
         anchor-free & cap. increase & seg loss & 5-frame & TTA & AP@0.7/0.8 \\
         \midrule
         \checkmark & - & - & - & - & 71.20 / 39.70 \\
         \checkmark & \checkmark & - & - & - & 74.28 / 42.91 \\
         \checkmark & \checkmark & \checkmark & - & - & 74.64 / 43.30 \\
         \checkmark & \checkmark & \checkmark & \checkmark & - & 76.34 / 45.57 \\ 
         \checkmark & \checkmark & \checkmark & \checkmark & \checkmark & 79.73 / 49.43 \\ 
         
         \bottomrule
    \end{tabular}
    \caption{\textbf{Ablation studies on the improvements to 3D detector MVF}~\cite{zhou2020end}. Metrics are 3D AP (L1) at IoU thresholds 0.7 and 0.8 for vehicles on the Waymo Open Dataset \emph{val} set.}
    \label{tab:detector_ablation}
\end{table}

\begin{table}[]
    \centering
    \small
    \begin{tabular}{c|c|c|c|c|c|c}
    \toprule
         \# frames & 1 & 2 & 3 & 4 & 5 & 10 \\
         \midrule
         AP@0.7 & 74.64 & 75.32 & 75.63 & 76.17 & 76.34 & 76.96\\
         AP@0.8 & 43.30 & 44.11 & 44.80 & 45.43 & 45.57 & 46.20\\
         \bottomrule
    \end{tabular}
    \caption{\textbf{Ablation studies on 3D detection AP \vs temporal contexts.} Metrics are 3D AP (L1) for vehicles on the Waymo Open Dataset \emph{val} set. We used the 5-frame model in 3D Auto Labeling.
    }
    \label{tab:detection_AP_vs_frames}
\end{table}

\begin{figure*}[t!]
    \centering
    \includegraphics[width=\linewidth]{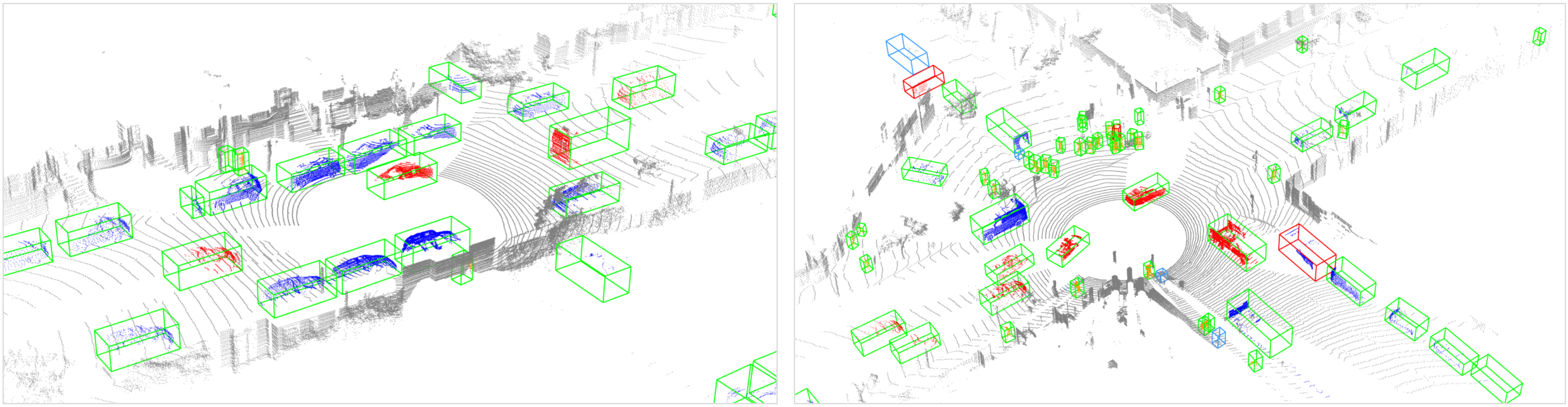}
    \caption{\textbf{Visualization of 3D auto labels on the Waymo Open Dataset \emph{val} set} (best viewed in color with zoom in). Object points are colored by object types with \textcolor{blue}{blue} for static vehicles, \textcolor{red}{red} for moving vehicles and \textcolor{YellowOrange}{orange} for pedestrians. Boxes are colored as: \textcolor{green}{green} for true positive detections, \textcolor{red}{red} for false positives and \textcolor{cyan}{cyan} for ground truth boxes in the cases of false negatives.}
    \label{fig:visualization}
\end{figure*}

\subsection{Analysis of Object Auto Labeling Models}
\label{sec:exp:analysis}


We evaluate the object auto labeling models using the box accuracy metric under two IoU thresholds 0.7 and 0.8 on the Waymo Open Dataset \emph{val} set. The predicted box is considered correct if its IoU with the ground truth is higher than the threshold. More analysis is in Sec.~\ref{sec:supp:analysis}.

\paragraph{Ablations of the static object auto labeling}
In table~\ref{tab:static_ablation} we can see the importance of the initial coordinate transform (to the box coordinate), and the foreground segmentation network in the first 3 rows. In the 4th and the 5th rows, we see the gains of using iterative box re-estimation and test time augmentation respectively.

\paragraph{Alternative designs of the dynamic object auto labeling}
Table~\ref{tab:dynamic_alternative} ablates the design of the dynamic object auto labeling model. For the align \& refine model, we use the multi-frame MVF++ detector boxes to ``align'' the object point clouds from the nearby frames ($[-2,+2]$) to the center frame. For each context frame, we transform the coordinate by aligning the center and heading of the context frame boxes to the center frame box. The model using un-aligned point clouds (in the center frame's coordinate, from $[-2,+2]$ context frames), second row, actually gets higher accuracy (second row) than the aligned one.
The model taking only the box sequence (third row) as input performs reasonably as well, by leveraging the trajectory shape and the box sizes. Our model jointly using the multi-frame object point clouds and the box sequences gets the best accuracy.

\begin{table}[t]
    \centering
    \small
    \begin{tabular}{c|c|c|c|c}
    \toprule
         transform & segmentation & iterative & tta & Acc@0.7/0.8 \\
         \midrule
         - & - & - & - & 78.82 / 50.90 \\
         \checkmark & - & - & - & 81.35 / 54.76 \\
         \checkmark & \checkmark & - & - & 81.37 / 55.67 \\ \midrule
         \checkmark & \checkmark &  \checkmark & - & 82.02 / 56.77\\
         \checkmark & \checkmark &  \checkmark & \checkmark & \textbf{82.28} / \textbf{56.92} \\
         \bottomrule
    \end{tabular}
    \caption{\textbf{Ablation studies of the static auto labeling model.} Metrics are the box accuracy at 3D IoU=0.7 and IoU=0.8 for vehicles in the Waymo Open Dataset \emph{val} set.}
    \label{tab:static_ablation}
\end{table}

\begin{table}[t]
    \centering
    \small
    \begin{tabular}{l|c}
    \toprule
        Method & Acc@0.7/0.8 \\
        \midrule
         Align \& refine&  83.33 / 60.69 \\
         Points only & 83.79 / 61.95 \\
         Box sequence only & 83.13 / 58.96 \\ 
         \midrule
         Points and box sequence joint & \textbf{85.67} / \textbf{65.77} \\ 
         \bottomrule
    \end{tabular}
    \caption{\textbf{Comparing with alternative designs of dynamic object auto labeling.}
    Metrics are box accuracy with 3D IoU thresholds 0.7 and 0.8 for vehicles on the Waymo Open Dataset \emph{val} set.
    }
    \label{tab:dynamic_alternative}
\end{table}

\paragraph{Effects of temporal context sizes for object auto labeling}
Table~\ref{tab:context_size} studies how the context frame sizes influence the box prediction accuracy. We also compare with our single-frame (S-MVF++) and multi-frame detectors (M-MVF++) to show extra gains the object auto labeling can bring. We can clearly see that using large temporal contexts improves the performance while using the entire object track (the last row) leads to the best performance. Note that for the static object model, we use the detector box with the highest score for the initial coordinate transform, which gives our auto labeling an advantage over frame-based method.


\begin{table}[t!]
    \centering
    \small
    \begin{tabular}{l|c|c|c}
         \toprule
         \multirow{2}{*}{Method} & \multirow{2}{*}{Context frames} & \emph{static} & \emph{dynamic} \\
         & & Acc@0.7/0.8 & Acc@0.7/0.8 \\
         \midrule
        S-MVF++ & $[-0,+0]$ & 67.17 / 36.61 & 80.07 / 57.71 \\
        M-MVF++ & $[-4,+0]$  & 73.96 / 43.56 & 82.21 / 59.52 \\
        \midrule
         \multirow{4}{*}{3DAL} & $[-0,+0]$ & 78.13 / 50.30 & 80.65 / 57.97 \\ 
         & $[-2,+2]$ & 79.60 / 52.52 & 84.34 / 63.60 \\
         & $[-5,+5]$ & 80.48 / 55.02 & 85.10 / 64.51 \\
         & all & \textbf{82.28} / \textbf{56.92} & \textbf{85.67} / \textbf{65.77} \\
         \bottomrule
    \end{tabular}
    \caption{\textbf{Effects of temporal context sizes for object auto labeling.} Metrics are the box accuracy at 3D IoU=0.7, 0.8 for vehicles in the WOD \emph{val} set. Dynamic vehicles have a higher accuracy because they are closer to the sensor than static ones.
    }
    \label{tab:context_size}
\end{table}



\subsection{Qualitative Analysis}
\label{sec:exp:visualization}

In Fig.~\ref{fig:visualization}, we visualize the auto labels for two representative scenes in autonomous driving: driving on a road with parked cars, and passing a busy intersection. Our model is able to accurately recognize vehicles and pedestrians in challenging cases with occlusions and very few points. The busy intersection scene also shows a few failure cases including false negatives of pedestrians in rare poses (sitting), false negatives of severely occluded objects and false positive for objects with similar geometry to cars. Those hard cases can potentially be solved with added camera information with multi-modal learning.


%% file: tex/conclusion.tex
In this work we have introduced 3D Auto Labeling, a state-of-the-art offboard 3D object detection solution using point cloud sequences as input. The pipeline leverages the long-term temporal data of objects in the 3D scene. Key to our success are our object-centric formulation, powerful offboard multi-frame detector and novel object auto labeling models.
Evaluated on the Waymo Open Dataset, our solution has shown significant gains over prior art onboard 3D detectors, especially with high standard metrics. A human label study has further shown the high quality of the auto labels reaching comparable performance as experienced humans. Moreover, the semi-supervised learning experiments have demonstrated the usefulness of the auto labels for student training in cases of low-label and unseen domains.

%% file: tex/supplementary.tex
\section{Overview}
In this document, we provide more details of models, experiments and show more analysis results. Sec.~\ref{sec:supp:eval} presents more evaluation results on the Waymo Open Dataset test set and shows how our offboard 3D detection can help domain adaptation and 3D tracking. Sec.~\ref{sec:supp:mvf} explains more details of MVF++ detectors. Sec.~\ref{sec:supp:tracker} and Sec.~\ref{sec:supp:motion} describe implementation details of our multi-object tracker and track-based motion state classifier respectively. Sec.~\ref{sec:supp:object_auto_labeling} covers network architectures, losses and training details of object auto labeling models. Sec.~\ref{sec:supp:human} describes the specifics of the human label study for 3D object detection and provides more statistics. Sec.~\ref{sec:supp:semi} provides more information about the semi-supervised learning experiments. Lastly Sec.~\ref{sec:supp:analysis} gives more analysis results supplementary to the main paper.

\section{More Evaluation Results}
\label{sec:supp:eval}

\subsection{3D Detection Results on the Test Set}
In Table.~\ref{tab:detection_test_set} we report detection results on the Waymo Open Dataset \emph{test} set comparing our pipeline with a few leading methods in the leaderboard~\cite{wod_3d_detection_website}. Note that our pipeline achieves the best results among Lidar-only methods. It also outperforms the HorizonLidar3D which uses both camera and Lidar input in the L1 metrics. We expect that adding camera input to our pipeline can further improve our pipeline in hard cases (L2).

\begin{table}[h]
    \centering
    \small
    \resizebox{\columnwidth}{!}{
    \begin{tabu}{l|c|cccc}
    \toprule
        Method & Sensor & AP L1 & APH L1 & AP L2 & APH L2 \\ \midrule
        PV-RCNN & L & 81.06 & 80.57 & 73.69 & 73.23 \\
        CenterPoint & L & 81.05 & 80.59 & 73.42 & 72.99 \\
        HorizonLidar3D & CL & 85.09 & 84.68 & \textbf{78.23} & \textbf{77.83} \\
        \midrule
        3DAL (ours) & L & \textbf{85.84} & \textbf{85.46} & 77.24 & 76.91 \\
    \bottomrule
    \end{tabu}
    }
    \caption{\textbf{3D detection AP on the Waymo Open Dataset main \emph{test} set for vehicles.} Evaluation results were obtained from submitting to the test server. For the sensor the `L` means Lidar-only; the `CL` means camera and Lidar. Note that our method peeks into the future for object-centric refinement, which is feasible in the offboard setting.}
    \label{tab:detection_test_set}
\end{table}

\subsection{Domain Adaptation Results}
In Table~\ref{tab:domain_adaptation} we report detection results in another domain and compare our 3D Auto Labeling (3DAL) pipeline with two baselines: the popular PointPillars~\cite{lang2019pointpillars} detector and our offboard multi-frame MVF++ detector. We see that our 3D Auto Labeling pipeline achieves significantly higher detection APs compared to the baselines (\textbf{32.56} higher 3D AP than the PointPillars and \textbf{8.03} higher 3D AP than the multi-frame MVF++), showing the strong generalization ability of our models. Compared to a few leading methods on the leaderboard~\cite{wod_domain_adaptation_website} our method also shows significant gains. These large gains are probably due to the temporal information aggregation, which compensates the lower point densities in the WOD domain adaptation set (collected in Kirkland with mostly rainy weather). 

\begin{table}[t]
    \centering
    \small
    \begin{tabular}{l|c|c|c|c}
    \toprule
        Method & 3D AP & 0-30m & 30-50m & 50+m \\ \midrule
        PointPillar & 45.48 & 74.02 & 36.49 & 14.94\\
        Multi-frame MVF++ & 70.01 & 86.54 & 67.72 & 43.25 \\
        \midrule
        PV-RCNN-DA & 71.40 & 90.00 & 66.45 & 45.92 \\
        CenterPoint & 67.04 & 86.62 & 60.95 & 38.59 \\
        HorizonLidar3D & 72.48 & 90.65 & 67.26 & 47.89 \\
        \midrule
        3DAL (ours) & \textbf{78.04} & \textbf{91.90} & \textbf{73.47} & \textbf{52.53} \\
    \bottomrule
    \end{tabular}
    \caption{\textbf{3D detection AP on the Waymo Open Dataset domain adaptation \emph{test} set for vehicles.} The PointPillar, MVF++ and 3DAL models were trained by us on the Waymo Open Dataset main \emph{train} set. Evaluation results were obtained from submitting to the test server. The PV-RCNN-DA, CenterPoint and HorizonLidar3D results are leading entries from the leaderboard~\cite{wod_domain_adaptation_website}.}
    \label{tab:domain_adaptation}
\end{table}

\subsection{3D Tracking Results}
In table~\ref{tab:tracking_results} we show how our improved box estimation from the offboard 3D Auto Labeling enhances the tracking performance, compared to using the boxes from the single-frame or multi-frame detectors. All methods used the same tracker (Sec.~\ref{sec:supp:tracker}). This reflects that in the tracking-by-detection paradigm, the localization accuracy plays an important role in determining tracking quality in terms of MOTA and MOTP.

\begin{table}[h]
    \centering
    \small
    \begin{tabular}{l|c|c}
         \toprule
         Method & MOTA$\uparrow$ & MOTP$\downarrow$ \\
         \midrule
         Single-frame MVF++ with KF & 52.20 & 17.08 \\
         Multi-frame MVF++ with KF & 61.92 & 16.31 \\
         \midrule
         3D Auto Labeling & \textbf{66.90} & \textbf{15.45} \\ 
         \bottomrule
    \end{tabular}
    \caption{\textbf{3D tracking results for vehicles on the Waymo Open Dataset \emph{val} set.} The metrics are L1 MOTA and MOTP for vehicles on the Waymo Open Dataset \emph{val} set. KF stands for using Kalman Filtering for the track state update. The arrows indicate whether the metric is better when it is higher (up-ward arrow) or is better when it is lower (down-ward arrow).}
    \label{tab:tracking_results}
\end{table}

\section{Implementation Details of the MVF++ Detectors}
\label{sec:supp:mvf}
\paragraph{Network Architecture}
\begin{figure*}[t]
    \centering
    \includegraphics[width=0.95\linewidth]{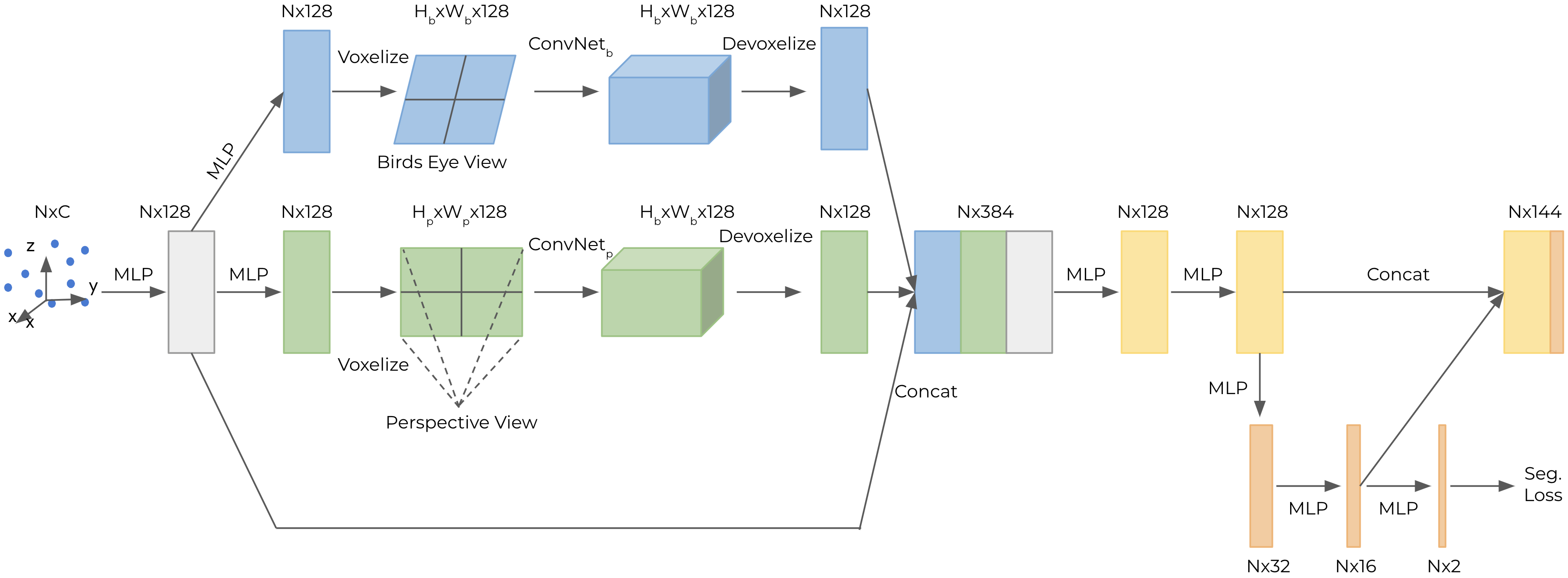}
    \caption{\textbf{Point-wise feature fusion network of MVF++.} Given an input point cloud encoding of shape $N \times C$, the network maps it to high-dimensional feature space and extracts contextual information from different views \ie the Bird's Eye View and the Perspective View. It fuses view-dependent features by concatenating information from three sources. The final output has shape $N \times 144$, as a result of concatenating dimension-reduced point features of shape $N \times 128$ with 3D segmentation features of shape $N \times 16$.} 
    \label{fig:mvfpp_pointfusion_pipeline}
\end{figure*}
Figure~\ref{fig:mvfpp_pointfusion_pipeline} illustrates the point-wise feature fusion network within the proposed MVF++. Given $C$-dimensional input encoding of $N$ points~\cite{zhou2020end}, the network first projects the points into a $128$-D feature space via a multi-layer perceptron (MLP), where shape information can be better described. The MLP is composed of a linear layer, a batch normalization(BN) layer and a rectified linear unit (ReLU) layer. Then it processes the features by two separate MLPs for view-dependent information extraction, \ie one for the Bird's Eye View and one for the Perspective View~\cite{zhou2020end}. Next, the network employs voxelization~\cite{zhou2020end} to transform view-dependent point features into the corresponding 2D feature maps, which are fed to view-dependent ConvNets (\ie ConvNet$_{b}$ and ConvNet$_{p}$) to further extract contextual information within an enlarged receptive field. Different from MVF~\cite{zhou2020end} using one ResNet~\cite{he2016deep} layer in obtaining each down-sampled feature maps, we increase the depth of ConvNet$_{b}$ and ConvNet$_{p}$ by applying one more ResNet block in each down-sampling branch. At the end of view-dependent processing, it applies devoxelization to transform the 2D feature map back to point-wise features. The model fuses point-wise features by concatenating three sources of information. To reduce computational complexity, it applies two MLPs consecutively, reducing the feature dimension to 128. For improving the discriminative capability of features, it introduces 3D segmentation auxiliary loss and augments the dimension-reduced features with segmentation features. The output of point-wise feature fusion network has shape $N \times 144$. 

Upon obtaining point-wise features, we voxelize them into a 2D feature map and employ a backbone network to generate detection results. Specifically, we adopt the same architecture as in~\cite{lang2019pointpillars,zhou2020end}. To further boost detection performance in the offboard setting, we replace each plain convolution layer with a ResNet~\cite{he2016deep} layer maintaining the same output feature dimension and feature map resolution. 

\paragraph{Loss Function}
We train MVF++ by minimizing a loss function, defined as $L = L_{\textrm{cls}} + w_1 L_{\textrm{centerness}} + w_2 L_{\textrm{reg}} + w_3 L_{\textrm{seg}}$. $L_{\textrm{cls}}$ and $L_{\textrm{centerness}}$ are focal loss and centerness loss as in~\cite{REF:tian2019fcos}. $L_{\textrm{reg}}$ represents Smooth L1 loss learning to regress x, y, z center locations, length, width, height and heading orientation at foreground pixels, as in~\cite{lang2019pointpillars,zhou2020end}. $L_{\textrm{seg}}$ is the auxiliary 3D segmentation loss for distinguishing foreground from background points (points are labeled as foreground/background if they lie inside/outside of a ground truth 3D box)~\cite{shi2018pointrcnn,qi2018frustum}. In our experiments, we set $w_1=1.0$, $w_2=2.0$, $w_3=1.0$. At inference time, the final score for ranking all detected boxes is computed as the multiplication of the classification score and the centerness score. By doing so, the centerness score can downplay the boxes far away from an object center and thus encourage non-maximum suppression (NMS) to yield high-quality boxes, as recommended in~\cite{REF:tian2019fcos}.  

\paragraph{Data Augmentation}
We perform three global augmentations that are applied to the LiDAR point cloud and ground truth boxes simultaneously~\cite{zhou2018voxelnet}. First, we apply random flip along the x axis, with probability 0.5. Then, we employ a random global rotation and scaling, where the rotation angle and the scaling factor are randomly drawn uniformly from $[-\pi/4,+\pi/4]$ and $[0.9, 1.1]$, respectively. Finally, we add a global translation noise to x, y, z drawn from $\mathcal{N}(0, 0.6)$.

\paragraph{Hyperparameters} For vehicles, we set voxel size to $[0.32, 0.32, 6.0]$m and detection range to $[-74.88, 74.88]$m along the X and Y axes and $[-2, 4]$m along Z axis, which results in a $468 \times 468$ 2D feature map in the Bird's Eye View. For pedestrians, we set voxel size to $[0.24, 0.24, 4.0]$m and detection range to $[-74.88, 74.88]$m along the X and Y axes and $[-1, 3]$m along Z axis, which corresponds to a $624 \times 624$ 2D feature map in the Bird's Eye View. During test-time augmentation, we set (IoU threshold, box score) to be (0.275, 0.5) for vehicles and (0.2, 0.5) for pedestrians, to trigger weighted box fusion~\cite{REF:wbf2019}.

\paragraph{Training} During training, we use the Adam optimizer~\cite{REF:Adam} and apply cosine decay to the learning rate. The initial learning rate is set to $1.33 \times 10^{-3}$ and ramps up to $3.0 \times 10^{-3}$ after 1000 warm-up steps. The training used 64 TPUs with a global batch size of 128 and finished after 43,000 steps.


\section{Implementation Details of the Tracker}
\label{sec:supp:tracker}
Our multi-object tracker is a similar implementation to~\cite{weng2019baseline}.  To reduce the impact of sensor ego-motion in tracking, we transformed all the boxes to the world coordinate for tracking. To reduce false positives, we also filter out all detections with scores less than 0.1 before the tracking. We used Bird's Eye View (BEV) boxes for detection and track association, using the Hungarian algorithm with an IoU threshold of 0.1. During the states update, the heading is handled specially as there can be flips and cyclic patterns. Before updating the heading state, we first adjust the detection heading to align with the track state heading -- if the angle difference is obtuse, we add $\pi$ to the detection angle before the update; we also average the angles in the cyclic space (\eg the average of 6 rad and 0.5 rad is 0.1084 rather than 3.25).

\section{Implementation Details of the Motion State Estimator}
\label{sec:supp:motion}
As we introduced in the main paper, we use the object track data for motion state estimation, which is much easier compared to classifying the static/non-static state from a single or a few frames. Note that we define an object as static only if it is stationary in the entire sequence. Specifically, we extract two heuristic-based features and fit a linear classifier to estimate the motion state. The two features are: the detection box centers' variance and the begin-to-end distance of the tracked boxes (the distance from the center of the first box of the track to the center of the last box of the track), with boxes all in the world coordinate. To ensure that the statistics are reliable we only consider tracks with at least 7 valid measurements. For tracks that are too short, we do not run the classification nor the auto labeling models. The boxes of those short tracks are merged directly to the final auto labels.

The ground truth motion states are computed from ground truth boxes with pre-defined thresholds of begin-to-end distance (1.0m) and max speed (1m/s). The thresholds are needed because there could be small drifts in sensor poses, such that the ground truth boxes in the world coordinate are not exactly the same for a static object.

For vehicles, such a simple linear model can achieve more than $99\%$ classification accuracy. The remaining rare error cases usually happen in short tracks with noisy detection boxes, or for objects that are heavily occluded or far away. For pedestrians, as most of them are moving and even the static ones tend to move their arms and heads, we consider all pedestrian tracks as dynamic.

\section{Details of the Object Auto Labeling Models}
\label{sec:supp:object_auto_labeling}

\subsection{Static Object Auto Labeling}
\paragraph{Network architecture.}
In the static object auto labeling model, the foreground segmentation is a PointNet~\cite{qi2017pointnet} segmentation network, where each point is firstly processed by an multi-layer perceptron (MLP) with 5 layers with output channel sizes of $64, 64, 64, 128, 1024$. For every layer of the MLP, we have batch normalization and ReLU. The $1024$-dim per point embeddings are pooled with a max pooling layer and concatenated with the output of the 2nd layer of the per-point MLP ($64$-dim). The concatenated $1088$-dim features are further processed by an MLP of 5 layers with output channel sizes $512,256,128,128,2$, where the last layer does not have non-linearity or batch normalization. The predicted foreground logit scores are used to classify each point as foreground or background. All the foreground points are extracted.

The box regression network is also a PointNet~\cite{qi2017pointnet} variant that takes the foreground points and outputs the 3D box parameters. It has a per-point MLP with output sizes of $128, 128, 256, 512$, a max pooling layer and a following MLP with output sizes $512, 256$ on the max pooled features. There is a final linear layer predicting the box parameters. We parameterize the boxes in a way similar to~\cite{qi2018frustum} as the box center regression ($3$-dim), the box heading regression and classification (to each of the heading bins) and the box size regression and classification (to each of the template sizes). For iterative refinement, we apply the same box regression network one more time on the foreground points transformed to the estimated box's coordinate. We found that if we use multi-frame MVF++ boxes, using shared weights for the two box regression networks works better than not sharing the weights; while if we use the single-frame MVF++, the cascaded design without sharing the weights works better. The numbers in the main paper are from the iterative model (shared weights).

For simplicity and higher generalizability of the model, we only used $XYZ$ coordinates of the points in the segmentation and box regression networks. Intensities and other point channels were not used. We have also tried to use the more powerful PointNet++~\cite{qi2017pointnetplusplus} models but did not see improvement compared to the PointNet-based models in this problem.

\paragraph{Losses.}
The model is trained with supervision of the segmentation masks and the ground truth 3D bounding boxes. For the segmentation, the sub-network predicts two scores for each point as foreground or background and is supervised with a cross-entropy loss $L_{seg}$. For the box regression, we implement a process similar to~\cite{qi2018frustum}, where each box regression network regresses the box by predicting its center $cx, cy, cz$, its size classes (among a few pre-defined template size classes) and residual sizes for each size class, as well as the heading bin class and a residual heading for each bin. We used $12$ heading bins (each bin account for 30 degrees) and $3$ size clusters: $(4.8, 1.8, 1.5), (10.0, 2.6, 3.2), (2.0, 1.0, 1.6)$, where the dimensions are length, width, height. The box regression loss is defined as $L_{\text{box}_i} = L_{\text{c-reg}_i} + w_1 L_{\text{s-cls}_i} + w_2 L_{\text{s-reg}_i} + w_3 L_{\text{h-cls}_i} + w_4 L_{\text{h-reg}_i}$ where $i \in \{1,2\}$ represents the cascade/iterative box estimation step. The total box regression loss is $L = L_{seg} + w (L_{\text{box}_1} + L_{\text{box}_2})$. The $w_i$ and $w$ are hyperparameter weights of the losses. Empirically, we use $w_1 = 0.1$, $w_2 = 2$, $w_3 = 0.1$, $w_4 = 2$ and $w = 10$.

\paragraph{Training and data augmentation.}
We train our models using the extracted object tracks (with the proposed multi-frame MVF++ model and our multi-object tracker) from the Waymo Open Dataset for each class type separately. Ground truth boxes are assigned to every frame of the track (frames with no matched ground truth are skipped). 

During training, for each static object track, we randomly select an initial box from the sequence. We also randomly sub-sample $\text{Uniform}[1, |S_j|]$ frames from all the visible frames $S_j$ of an object $j$. This naturally leads to a data augmentation effect. Note that at test time we always select the initial box with the highest score and use all frames. The merged points are randomly sub-sampled to $4,096$ points and randomly flipped along the $X, Y$ axes with $50\%$ chance respectively and randomly rotated around the up-axis ($Z$) by $\text{Uniform}[-10, 10]$ degrees.
To increase the data quantity, we also turn the dynamic object track data to pseudo static track. To achieve that, we use the ground truth object boxes to align the dynamic object points to a specific frame's ground truth box coordinate. This increases the number of object tracks of vehicles by 30\%.

In total, we have extracted around 50K (vehicle) object tracks for training (including the augmented ones from dynamic objects) and around 10K object tracks for validation (static only). We trained the model using the Adam optimizer with a batch size of 32 and an initial learning rate of 0.001. The learning rate was decayed by 10X at the 60th, 100th and 140th epochs. The model was trained with 180 epochs in total, which took around 20 hours with a V100 GPU.

\subsection{Dynamic Object Auto Labeling}
\paragraph{Network architecture.}
For the foreground segmentation network, we adopt a similar architecture as that for the static auto labeling model except that the input points have one more channel besides the $XYZ$ coordinate, the time encoding channel. The temporal encoding is $0$ for points from the current frame, $-0.1r$ for the $r$-th frame prior to the current frame and $+0.1r$ for the $r$-th frame after the current frame. In our implementation we take 5 frames of object points with each frame's points subsampled to $1,024$ points, so in total there are $5,120$ points input to the segmentation network. The point sequence encoder network takes the segmented foreground points and uses a PointNet~\cite{qi2017pointnet}-like architecture with a per-point MLP of output sizes $64,128,256,512$, a max-pooling layer and another MLP with output sizes $512, 256$ on the max-pooled features. The output is a $256$-dim feature vector.

For the box sequence encoder network, we consider each box (in the center frame's box coordinate) as a parameterized point with channels of box center (cx, cy, cz), box size (length, width, height), box heading $\theta$ and a temporal encoding. We use nearly the entire box sequence (setting $s$ in the main paper to 50, leading to a sequence length of 101).

The box sequence can be considered as a point cloud and processed by another PointNet. Note, such a sequence can also be processed by a 1D ConvNet, or be concatenated and processed by a fully connected network, or we can even use a graph neural network. Through empirical study we found using a PointNet to encode the box sequence feature is both effective (compared with ConvNet and fully connected layers) and simple (compared with graph neural networks). The box sequence encoding PointNet has a per-point MLP with output sizes $64, 64, 128, 512$, a max-pooling layer and another MLP with output sizes $128, 128$ on the max-pooled features. The final output is a $128$-dim feature vector, which we call the trajectory embedding.

The point embedding and the trajectory embedding are concatenated and passed through a final box regression network, which is a MLP with two layers with output sizes $128, 128$ and a linear layer to regress the box parameters (similar to that of the static object auto labeling model).

To encourage contributions from both branches, we follow~\cite{wang2019makes} and also pass the trajectory embedding and the object embedding to two additional box regression sub-networks to predict boxes independently. The sub-networks have the same structure as the one for the joint-embedding, but with non-shared weights.

\paragraph{Losses.}
Similar to the static auto labeling model, we have two types of loss, the segmentation loss and the box regression loss. The box regression outputs are defined in the same way as that for the static objects. We used $12$ heading bins (each bin account for 30 degrees) and the same size clusters as those for the static vehicle auto labeling. For pedestrians we use a single size cluster: $(0.9, 0.9, 1.7)$ of length, width, height. The final loss is $L = L_{\text{seg}} + v_1 L_\text{box-traj} + v_2 L_\text{box-obj-pc} + v_3 L_\text{box-joint}$ where we have three box losses from the trajectory head, the object point cloud head and the joint head respectively. The $v_i$, $i=1,2,3$ are the weights for the loss terms to achieve a balanced learning of the three types of embeddings. Empirically, we use $v_1 = 0.3, v_2 = 0.3, v_3 = 0.4$.

\paragraph{Training and data augmentation.}
During training, we randomly select the center frame from each dynamic object track. If the context size is less than the required sequence length $2r+1$ or $2s+1$, or when the frames in the sequence are not consecutive (\eg the object is occluded for a few frames), we use placeholder points and boxes (all zeros) for the empty frames. As there may be tracking errors, we match our object track with ground truth tracks and avoid training on the ones with switched track IDs.

As to augmentation, both points and boxes are randomly flipped along the $X$ and $Y$ axis with a $50\%$ chance and randomly rotated around the $Z$ axis by $\text{Uniform}[-10, 10]$ degrees. We also add a light random shift and a random scaling to the point clouds. Point cloud from each frame is also randomly sampled to $1,024$ points from the full point cloud observed.

 We train vehicle and pedestrian models separately. For vehicles we extracted around 15.7K dynamic tracks for training and 3K for validation. For pedestrians, we extracted around 22.9K dynamic tracks for training and around 5.1K for validation. We train the model using the Adam optimizer with batch size 32 and an initial learning rate 0.001. The learning rate is decayed by 10 times at the 180th, 300th and 420th epochs. The model is trained with 500 epochs in total, which takes 1-2 days with a V100 GPU.

\section{Details of the Human Label Study}
\label{sec:supp:human}

We randomly selected 5 sequences from the Waymo Open Dataset \emph{val} set as listed in Table~\ref{tab:human_label_segments} to run the human label study. The 15 labeling tasks (3 sets of re-labels for each run segment) involved 12 labelers with experiences in labeling 3D Lidar point clouds. In total we collected around 2.3K labels (one label for one object track) for the 3 repeated labelings.

\begin{table}[]
    \centering
    \begin{tabular}{l}
    \toprule
         segment-17703234244970638241\_220\_000\_240\_000 \\
         segment-15611747084548773814\_3740\_000\_3760\_000 \\
         segment-11660186733224028707\_420\_000\_440\_000 \\
         segment-1024360143612057520\_3580\_000\_3600\_000 \\
         segment-6491418762940479413\_6520\_000\_6540\_000 \\
         \bottomrule
    \end{tabular}
    \caption{\textbf{Sequence (run segment) list for the human label study.} The sequences are all from the Waymo Open Dataset \emph{val} set.}
    \label{tab:human_label_segments}
\end{table}


\paragraph{How consistent are human labels?} Auxiliary to the AP results in the main paper, we also analyze the IoUs between human labels. We found that even for humans, 3D bounding box labeling can be challenging as the input point clouds are often partial and occluded. To understand how consistent human labels are, we compare labels from one labeler with the labels from the other and measure the 3D box consistency by their IoUs. Since we already have the verified public ground truth, we can compare the 3 sets of labels with the public WOD ground truth and get the average box IoU for all objects that are matched (due to occlusions, some objects may be labeled or not labeled by a specific labeler). Specifically, for human boxes that we cannot find a ground truth box with more than 0.03 BEV IoU overlap (false positive or false negative), they were ignored and not counted in the computation.

The statistics are summarized in Table~\ref{tab:human_mean_iou}. Surprisingly, human labels do not have the consistency one may expect (\eg 95\% IoU). Due to the inherent uncertainty of the problem, even humans can only achieve around 81\% 3D IoU or around 88\% BEV IoU in their box consistency. As we break down the numbers by distance we see, intuitively, that nearby objects have a significantly higher mean IoU as they have more visible points and more complete viewpoints. The BEV 2D IoU is also higher than the 3D IoU as we do not require the correct height estimation in the BEV box, which simplifies the problem.

To have a rough understanding of how the boxes generated by our 3D Auto Labeling pipeline compare with human labels, we compute the average IoU of auto labels with the WOD ground truth. Note that those numbers are not directly comparable to the average human IoUs as they cover different sets of objects (due to the false positives and false negatives). However, it still gives us an understanding that the auto labels are already on par in quality to human labels.

\section{More Details about the Semi-supervised Learning Experiment}
\label{sec:supp:semi}
In the semi-supervised experiments, we use an onboard single-frame \emph{MVF++} detector as the student. We train all networks with an effective batch size of 256 scenes per iteration. The training schedule starts with a warmup period where the learning rate is gradually increased to 0.03 in 1000 iterations. Afterward, we use a cosine decay learning rate schedule to drop the learning rate from 0.03 to 0. 

For the intra-domain semi-supervised learning, we randomly select 10\% of the sequences (around 15K frames from 79 sequences) to train the 3DAL pipeline which gets an AP of 78.11\% on the \emph{validation} set. Then, the 3DAL annotates the rest of the training set (around 142K frames from 719 sequences). Finally, the student is trained on the union of these sets (798 sequences). We train the models for a total of 43K iterations. 

For the cross-domain semi-supervised learning, we first train 3DAL on the regular Waymo Open \emph{training} set. Since the domain adaptation validation set is relatively small (\ie only contains 20 sequences), we submit the results to the submission server and report on the domain adaptation \emph{test} set (containing 100 sequences). The 3DAL gets an AP of 78.0\% on the domain adaptation \emph{test} set without using any data from that domain. Then, we use the trained pipeline to annotate the domain adaptation \emph{training} and \emph{unlabeled} sets of the Waymo Open Dataset. Finally, the student is trained on the union of the regular \emph{training} set annotated by humans, and domain adaptation \emph{training+unlabeled} sets annotated by 3DAL. Since the data used for training the student is around 2X larger compared to the intra-domain experiment, we also increase the training iterations to 80K.

\begin{table}[]
    \centering
    \small
    \begin{tabular}{c|c|p{0.7cm}|p{0.7cm}|p{0.7cm}|p{0.7cm}}
    \toprule
         IoU type & Label type & all & 0-30m & 30-50m & 50m+ \\ \midrule
         \multirow{2}{*}{valid boxes} & human & 25,641 & 11,543 & 7,963 & 6,135  \\
         & auto & 24,146 & 11,360 & 7,448 & 5,338 \\
         \midrule
         \multirow{2}{*}{3D mIoU} & human & 80.92 & 85.78 & 80.29 & 72.59 \\
         & auto & 80.29 & 84.04 & 77.45 & 76.28 \\
         \midrule
         \multirow{2}{*}{BEV mIoU} & human & 87.98 & 91.26 & 87.31 & 82.68 \\ 
         & auto & 87.50 & 90.36 & 85.09 & 84.78 \\
         \bottomrule
    \end{tabular}
    \caption{\textbf{The mean IoU of human labels and auto labels compared with the Waymo Open Dataset ground truth for vehicles.} Note that since different labels (human or machine) annotate different number of objects for each frame, those numbers are not directly comparable. They are summarized here for a reference. For a more fair comparison between human and auto labels, see the Average Precision comparison table in the main table. We only evaluate using ground truth boxes with at least one point in it and only evaluate boxes that have a BEV IoU larger than 0.03 with any ground truth box.}
    \label{tab:human_mean_iou}
\end{table}


\section{More Analysis Experiments for Object Auto Labeling}
\label{sec:supp:analysis}

In this section, we provide more analysis results auxiliary to the main paper.

\paragraph{Effects of key frame selection for static object auto labeling.}
Table~\ref{tab:static_frame_selection} compares the effects of using different initial boxes (from the detectors) for the model (for the coordinate transform before foreground segmentation): a uniformly chosen random box, the average box and the box with the highest score. We also show the box accuracy of the detectors as a reference (\ie the accuracy of those initial boxes).

Choosing a uniformly random box from the sequence is equivalent to the setting of a frame-centric approach. As for the detector baselines (row 1 and row 2), it directly evaluates the average accuracy of the detector boxes. As for the auto labeling, it means the box estimation is running for every frame, similar to a two-stage refinement step in two-stage detectors. We see that such a frame-centric box estimation achieves the most unfavorable results, as it is not able to leverage the best viewpoint in the sequence (in the object-centric way).

In the average box setting, we average all the boxes from the sequence (in the world coordinate) and use the averaged box for the transformation. For the highest score setting, we select the box with the highest confidence score as the initial box, which is similar to choosing the \emph{best} viewpoint of the object. We see that the strategy to choose the initial box has a great impact and can cause a 4.46 Acc@0.8 difference for the auto labeling model (the highest score box \vs a random box).

\begin{table}[]
    \centering
    \small
    \begin{tabular}{c|c|c}
    \toprule
         Model &  detector box & Acc@0.7/0.8 \\
         \midrule
         Single-frame MVF++ & random & 67.17 / 36.61 \\
         \midrule
         \multirow{3}{*}{Multi-frame MVF++} & random & 73.96 / 43.56 \\
          & average & 79.29 / 48.67 \\
          & highest score & 78.67 / 52.42 \\
         \midrule
         \multirow{3}{*}{Auto labeling model} & random & 79.66 / 52.46\\
          & average & 81.22 / 53.96 \\
          & highest score & \textbf{82.28} / \textbf{56.92} \\
         \bottomrule
    \end{tabular}
    \caption{\textbf{Effects of initial box selection in static object auto labeling.} Numbers are averaged over 3 runs for the random boxes.}
    \label{tab:static_frame_selection}
\end{table}

\begin{table}[b]
    \centering
    \small
    \begin{tabular}{c|c|c}
         \toprule
         ref frame & context frames & Acc@0.7/0.8 \\
         \midrule
         \multirow{2}{*}{all highest score} & $[-0,+0]$ & 78.13 / 50.30 \\
         & all & \textbf{82.28} / \textbf{56.92} \\
         \midrule
         past highest score & all history & 77.56 / 49.21\\
         \bottomrule
    \end{tabular}
    \caption{\textbf{Effects of temporal contexts for static object auto labeling.} Note that for a causal model, we cannot output a single box for a static object -- we have to output the best estimation for every frame using the current frame and the history frames. We compute the average accuracy across all frames for the causal case.}
    \label{tab:static_context_size}
\end{table}

\begin{table}[b]
    \centering
    \small
    \begin{tabular}{c|c|c}
    \toprule
         Point cloud context & Box context & Acc@0.7/0.8 \\
         \midrule
         $[-2,+2]$ & all & \textbf{85.67} / \textbf{65.77} \\
         \midrule
         \multirow{1}{*}{$[-4,0]$} & all history & 84.30 / 62.68\\
         \bottomrule
    \end{tabular}
    \caption{\textbf{Effects of temporal contexts for dynamic object auto labeling.} For causal models, we only use the causal point and box sequence input.}
    \label{tab:dynamic_context_size}
\end{table}

\paragraph{Causal model performance.}
Table~\ref{tab:static_context_size} compares non-causal models and causal models (for static object auto labeling). The causal model was trained using the causal input (the last row) and only used causal input for inference at every frame. We see the causal model has a relatively lower accuracy compared to non-causal ones, probably due to two reasons. First, it has limited contexts especially for the beginning frames of the track. Second, the pool of initial boxes are much more restricted if the input has to be causal. For non-causal models, we can select the frame with the highest confidence as the key frame and use the box from that frame for the initial coordinate transform. However, for the causal model, it can only select the highest confidence box from the \emph{history} frames, which are not necessarily the well visible ones.

As the causal model's accuracy is even inferior to the one that just uses a single frame's points for refinement (the first row in Table~\ref{tab:static_context_size}), the ability to select the best key frame weighs more than the added points from a few history frames. Note that the performance is still better than the detector boxes without refinement (row 2 in Table~\ref{tab:static_frame_selection}) Such results indicate the benefit of having a non-causal model for the offboard 3D detection.

Table~\ref{tab:dynamic_context_size} reports a similar study for dynamic object auto labeling. We also see that the causal model's performance is inferior to the non-causal one, although it still improves upon the raw detector accuracy (Table 8 in the main paper row 2, where the multi-frame MVF++ gets 82.21 / 59.52 accuracy).

\begin{table}[]
    \centering
    \small
\begin{tabular}{cc|cc} 
\toprule
\multicolumn{2}{c|}{Static}        & \multicolumn{2}{c}{Dynamic}         \\
Aug.        & Acc.@0.7/0.8         & Aug.        & Acc@0.7/0.8           \\ 
\midrule
All         & \textbf{82.28/56.92} & All         & 85.67/\textbf{65.77}  \\
$-$D2S      & 81.72/55.96          & $-$Shift    & 85.15/65.23           \\
$-$FlipX    & 81.42/55.98          & $-$Scale    & 85.15/65.91           \\
$-$FlipY    & 81.50/55.49          & $-$FlipY    & \textbf{85.76}/63.66  \\
$-$RotateZ  & 81.72/56.52          & $-$RotateZ  & 84.94/64.20           \\
\bottomrule
\end{tabular}
    \caption{\textbf{Ablations of data augmentation.} We use different data augmentations for static objects and dynamic objects. ``All'' means all the augmentations are used. ``$-$X'' means removing a specific augmentation, ``X'' from the augmentation set. ``D2S'' represents the dynamic-to-static augmentation. Best results in each column are in bold.}
    \label{tab:data_aug}
\end{table}


\begin{table}[]
    \centering
    \small
\begin{tabular}{lc|cc|cc} 
\toprule
                            &         & \multicolumn{2}{c|}{3D AP} & \multicolumn{2}{c}{BEV AP}         \\
\multicolumn{2}{c|}{Tracker}          & MOT   & GT                & MOT             & GT               \\ 
\midrule
\multirow{2}{*}{Vehicle}    & IoU=0.7 & 84.50 & 85.77    & 93.30           & 96.74   \\
                            & IoU=0.8 & 57.82 & 58.81    & 84.88           & 86.18   \\ 
\midrule
\multirow{2}{*}{Pedestrian} & IoU=0.5 & 82.88 & 83.02    & 86.32  & 86.24            \\
                            & IoU=0.6 & 63.69 & 64.80    & 75.60           & 75.65   \\
\bottomrule
\end{tabular}
    \caption{\textbf{Effects of the tracking accuracy.} ``MOT'' stands for Multi Object Tracker. ``GT'' represents Ground Truth Tracker where the ground truth boxes are used. Best results of each comparable pairs are in bold.}
    \label{tab:different_tracker}
\end{table}

\paragraph{Ablations of data augmentation.}
Table~\ref{tab:data_aug} compares the performance of our auto labeling pipeline when trained with different data augmentations.
The most accurate models are consistently trained with all the proposed augmentations. 
For static objects, all the augmentations contribute similarly in terms of accuracy, while for dynamic objects, random rotation around Z-axis appears to be the most critical.
%


\paragraph{Effects of the tracking accuracy.}
To study how tracking (association) accuracy affects our offboard 3D detection, we compare results using our Kalman filter tracker and an ``oracle'' tracker (they share the detection and object auto labeling models, just the tracker is different). For the ``oracle'' tracker, we associate detector boxes using the ground truth boxes. Specifically, for each detector box, we find its closest ground truth box and assign the ground truth box's object ID to it.
In Table.~\ref{tab:different_tracker}, we observe that better performances can be obtained when a more reliable tracker is used, although the difference is subtle.
In particular, the ``oracle'' tracker introduces more improvement for vehicles than pedestrians.
This difference can imply that, for pedestrians, there is more space for improvement in detecting targets than associating detected boxes.

\begin{table}[t]
    \centering
    \small
    \begin{tabular}{c|c|c|c|c}
    \toprule
         Motion & \multicolumn{2}{c}{3D AP} & \multicolumn{2}{c}{BEV AP} \\
         State & IoU=0.7 & IoU = 0.8 & IoU = 0.7 & IoU = 0.8 \\ \midrule
         Pred & 84.50 & 57.82 & 93.30 & 84.88 \\ \midrule
         GT & 84.98 & 57.95 & 93.36 & 85.13 \\
         \bottomrule
    \end{tabular}
    \caption{\textbf{Effects of the motion state estimation on the offboard 3D detection.} The metric is AP for vehicles on the Waymo Open Dataset \emph{val} set. ``Pred'' means we are classifying the motion state (static or not) using our linear classifier. ``GT'' means we are using the ground truth boxes to classify the motion state.}
    \label{tab:motion_state}
\end{table}

\paragraph{Effects of the motion state estimation accuracy.}
In Table~\ref{tab:motion_state} we study how motion state classification accuracy affects the offboard 3D detection AP. We replace the motion state classifier (``Pred'') with a new one using the ground truth boxes (``GT'') for classification and see how much AP improvements it can bring us. We see that while there are some gains, they are not significant. This is understandable as our linear classifier can already achieve a 99\%+ accuracy.


\paragraph{Inference speed.}
Processing a 20-second sequence (200 frames with the 10Hz sensor input) using a V100 GPU, the detector takes the majority time (around 15 minutes) due to the multi-frame input and test-time augmentation. The tracking takes around 3s and the object-centric refinement takes around 25s, which is 28s in total, 0.14s per frame, or a 3\% extra time over the detection. In the offboard setting, we can run detection or the refinement steps in parallel to further reduce the processing latency.

%% file: main.bbl
\begin{thebibliography}{10}\itemsep=-1pt

\bibitem{wod_3d_detection_website}
Waymo open dataset: 3d detection challenge.
\newblock \url{https://waymo.com/open/challenges/3d-detection/}.
\newblock Accessed: 2021-01-25.

\bibitem{wod_domain_adaptation_website}
Waymo open dataset: Domain adaptation challenge.
\newblock \url{https://waymo.com/open/challenges/domain-adaptation/}.
\newblock Accessed: 2021-01-25.

\bibitem{acuna2018efficient}
David Acuna, Huan Ling, Amlan Kar, and Sanja Fidler.
\newblock Efficient interactive annotation of segmentation datasets with
  polygon-rnn++.
\newblock In {\em Proceedings of the IEEE conference on Computer Vision and
  Pattern Recognition}, pages 859--868, 2018.

\bibitem{behley2019semantickitti}
Jens Behley, Martin Garbade, Andres Milioto, Jan Quenzel, Sven Behnke, Cyrill
  Stachniss, and Jurgen Gall.
\newblock Semantickitti: A dataset for semantic scene understanding of lidar
  sequences.
\newblock In {\em Proceedings of the IEEE International Conference on Computer
  Vision}, pages 9297--9307, 2019.

\bibitem{REF:bewley2020range}
Alex Bewley, Pei Sun, Thomas Mensink, Dragomir Anguelov, and Cristian
  Sminchisescu.
\newblock Range conditioned dilated convolutions for scale invariant 3d object
  detection, 2020.

\bibitem{cai2018cascade}
Zhaowei Cai and Nuno Vasconcelos.
\newblock Cascade r-cnn: Delving into high quality object detection.
\newblock In {\em CVPR}, 2018.

\bibitem{castrejon2017annotating}
Lluis Castrejon, Kaustav Kundu, Raquel Urtasun, and Sanja Fidler.
\newblock Annotating object instances with a polygon-rnn.
\newblock In {\em Proceedings of the IEEE conference on computer vision and
  pattern recognition}, pages 5230--5238, 2017.

\bibitem{cvpr17chen}
Xiaozhi Chen, Huimin Ma, Ji Wan, Bo Li, and Tian Xia.
\newblock Multi-view 3d object detection network for autonomous driving.
\newblock In {\em CVPR}, 2017.

\bibitem{REF:FastPointRCNN_Jiaya_ICCV2019}
Y. {Chen}, S. {Liu}, X. {Shen}, and J. {Jia}.
\newblock Fast point r-cnn.
\newblock In {\em 2019 IEEE/CVF International Conference on Computer Vision
  (ICCV)}, pages 9774--9783, 2019.

\bibitem{deng2009imagenet}
Jia Deng, Wei Dong, Richard Socher, Li-Jia Li, Kai Li, and Li Fei-Fei.
\newblock Imagenet: A large-scale hierarchical image database.
\newblock In {\em CVPR}. IEEE, 2009.

\bibitem{deshmukh1996benchmarking}
Neeraj Deshmukh, Richard~Jennings Duncan, Aravind Ganapathiraju, and Joseph
  Picone.
\newblock Benchmarking human performance for continuous speech recognition.
\newblock In {\em Proceeding of Fourth International Conference on Spoken
  Language Processing. ICSLP'96}, volume~4, pages 2486--2489. IEEE, 1996.

\bibitem{REF:Vote3Deep_ICRA2017}
M. {Engelcke}, D. {Rao}, D.~Z. {Wang}, C.~H. {Tong}, and I. {Posner}.
\newblock Vote3deep: Fast object detection in 3d point clouds using efficient
  convolutional neural networks.
\newblock In {\em 2017 IEEE International Conference on Robotics and Automation
  (ICRA)}, pages 1355--1361, May 2017.

\bibitem{feng2019deep}
Di Feng, Xiao Wei, Lars Rosenbaum, Atsuto Maki, and Klaus Dietmayer.
\newblock Deep active learning for efficient training of a lidar 3d object
  detector.
\newblock In {\em 2019 IEEE Intelligent Vehicles Symposium (IV)}, pages
  667--674. IEEE, 2019.

\bibitem{REF:AFDET_CVPRW2020}
Runzhou Ge, Zhuangzhuang Ding, Yihan Hu, Yu Wang, Sijia Chen, Li Huang, and
  Yuan Li.
\newblock Afdet: Anchor free one stage 3d object detection, 2020.

\bibitem{gu2019hplflownet}
Xiuye Gu, Yijie Wang, Chongruo Wu, Yong~Jae Lee, and Panqu Wang.
\newblock Hplflownet: Hierarchical permutohedral lattice flownet for scene flow
  estimation on large-scale point clouds.
\newblock In {\em Proceedings of the IEEE Conference on Computer Vision and
  Pattern Recognition}, pages 3254--3263, 2019.

\bibitem{REF:SA_SSD_He_2020_CVPR}
Chenhang He, Hui Zeng, Jianqiang Huang, Xian-Sheng Hua, and Lei Zhang.
\newblock Structure aware single-stage 3d object detection from point cloud.
\newblock In {\em Proceedings of the IEEE/CVF Conference on Computer Vision and
  Pattern Recognition (CVPR)}, June 2020.

\bibitem{he2016deep}
Kaiming He, Xiangyu Zhang, Shaoqing Ren, and Jian Sun.
\newblock Deep residual learning for image recognition.
\newblock In {\em CVPR}, 2016.

\bibitem{REF:you_see_Hu_2020_CVPR}
Peiyun Hu, Jason Ziglar, David Held, and Deva Ramanan.
\newblock What you see is what you get: Exploiting visibility for 3d object
  detection.
\newblock In {\em Proceedings of the IEEE/CVF Conference on Computer Vision and
  Pattern Recognition (CVPR)}, June 2020.

\bibitem{REF:ConvLSTM_ECCV2020}
Rui Huang, Wanyue Zhang, Abhijit Kundu, Caroline Pantofaru, David~A. Ross,
  Thomas~A. Funkhouser, and Alireza Fathi.
\newblock An {LSTM} approach to temporal 3d object detection in lidar point
  clouds.
\newblock {\em CoRR}, 2020.

\bibitem{iscen2019label}
Ahmet Iscen, Giorgos Tolias, Yannis Avrithis, and Ondrej Chum.
\newblock Label propagation for deep semi-supervised learning.
\newblock In {\em Proceedings of the IEEE conference on computer vision and
  pattern recognition}, pages 5070--5079, 2019.

\bibitem{REF:Adam}
Diederik~P. Kingma and Jimmy Ba.
\newblock Adam: A method for stochastic optimization.
\newblock {\em CoRR}, 2014.

\bibitem{REF:AlexNet:2017}
Alex Krizhevsky, Ilya Sutskever, and Geoffrey~E. Hinton.
\newblock Imagenet classification with deep convolutional neural networks.
\newblock {\em Commun. ACM}, 60(6):84--90, May 2017.

\bibitem{REF:ku2018joint}
Jason Ku, Melissa Mozifian, Jungwook Lee, Ali Harakeh, and Steven~L Waslander.
\newblock Joint 3d proposal generation and object detection from view
  aggregation.
\newblock In {\em 2018 IEEE/RSJ International Conference on Intelligent Robots
  and Systems (IROS)}, pages 1--8. IEEE, 2018.

\bibitem{lang2019pointpillars}
Alex~H Lang, Sourabh Vora, Holger Caesar, Lubing Zhou, Jiong Yang, and Oscar
  Beijbom.
\newblock Pointpillars: Fast encoders for object detection from point clouds.
\newblock In {\em CVPR}, 2019.

\bibitem{lee2013pseudo}
Dong-Hyun Lee.
\newblock Pseudo-label: The simple and efficient semi-supervised learning
  method for deep neural networks.
\newblock In {\em Workshop on challenges in representation learning, ICML},
  volume~3, 2013.

\bibitem{lee2018leveraging}
Jungwook Lee, Sean Walsh, Ali Harakeh, and Steven~L Waslander.
\newblock Leveraging pre-trained 3d object detection models for fast ground
  truth generation.
\newblock In {\em 2018 21st International Conference on Intelligent
  Transportation Systems (ITSC)}, pages 2504--2510. IEEE, 2018.

\bibitem{REF:3DFCN_RSJ2017}
B. {Li}.
\newblock 3d fully convolutional network for vehicle detection in point cloud.
\newblock In {\em 2017 IEEE/RSJ International Conference on Intelligent Robots
  and Systems (IROS)}, pages 1513--1518, Sep. 2017.

\bibitem{REF:VeloFCN2016}
Bo Li, Tianlei Zhang, and Tian Xia.
\newblock Vehicle detection from 3d lidar using fully convolutional network.
\newblock In {\em RSS 2016}, 2016.

\bibitem{li2020joint}
Peiliang Li, Jieqi Shi, and Shaojie Shen.
\newblock Joint spatial-temporal optimization for stereo 3d object tracking.
\newblock In {\em Proceedings of the IEEE/CVF Conference on Computer Vision and
  Pattern Recognition}, pages 6877--6886, 2020.

\bibitem{REF:Multi_task_multisensor_fusion_CVPR2019}
M. {Liang}, B. {Yang}, Y. {Chen}, R. {Hu}, and R. {Urtasun}.
\newblock Multi-task multi-sensor fusion for 3d object detection.
\newblock In {\em 2019 IEEE/CVF Conference on Computer Vision and Pattern
  Recognition (CVPR)}, pages 7337--7345, 2019.

\bibitem{REF:ContFuse_ECCV2018}
Ming Liang, Bin Yang, Shenlong Wang, and Raquel Urtasun.
\newblock Deep continuous fusion for multi-sensor 3d object detection.
\newblock In {\em ECCV}, 2018.

\bibitem{ling2019fast}
Huan Ling, Jun Gao, Amlan Kar, Wenzheng Chen, and Sanja Fidler.
\newblock Fast interactive object annotation with curve-gcn.
\newblock In {\em Proceedings of the IEEE Conference on Computer Vision and
  Pattern Recognition}, pages 5257--5266, 2019.

\bibitem{lippmann1997speech}
Richard~P Lippmann.
\newblock Speech recognition by machines and humans.
\newblock {\em Speech communication}, 22(1):1--15, 1997.

\bibitem{liu2019flownet3d}
Xingyu Liu, Charles~R Qi, and Leonidas~J Guibas.
\newblock Flownet3d: Learning scene flow in 3d point clouds.
\newblock In {\em Proceedings of the IEEE Conference on Computer Vision and
  Pattern Recognition}, pages 529--537, 2019.

\bibitem{liu2019meteornet}
Xingyu Liu, Mengyuan Yan, and Jeannette Bohg.
\newblock Meteornet: Deep learning on dynamic 3d point cloud sequences.
\newblock In {\em Proceedings of the IEEE International Conference on Computer
  Vision}, pages 9246--9255, 2019.

\bibitem{REF:FaF_Luo_2018_CVPR}
Wenjie Luo, Bin Yang, and Raquel Urtasun.
\newblock Fast and furious: Real time end-to-end 3d detection, tracking and
  motion forecasting with a single convolutional net.
\newblock In {\em Proceedings of the IEEE Conference on Computer Vision and
  Pattern Recognition (CVPR)}, June 2018.

\bibitem{meng2020weakly}
Qinghao Meng, Wenguan Wang, Tianfei Zhou, Jianbing Shen, Luc Van~Gool, and
  Dengxin Dai.
\newblock Weakly supervised 3d object detection from lidar point cloud.
\newblock {\em arXiv preprint arXiv:2007.11901}, 2020.

\bibitem{REF:LaserNet++_CVPRW2019}
G.~P. {Meyer}, J. {Charland}, D. {Hegde}, A. {Laddha}, and C.
  {Vallespi-Gonzalez}.
\newblock Sensor fusion for joint 3d object detection and semantic
  segmentation.
\newblock In {\em 2019 IEEE/CVF Conference on Computer Vision and Pattern
  Recognition Workshops (CVPRW)}, pages 1230--1237, 2019.

\bibitem{REF:lasernet_CVPR2019}
Gregory~P. Meyer, Ankit Laddha, Eric Kee, Carlos Vallespi-Gonzalez, and Carl~K.
  Wellington.
\newblock {LaserNet}: An efficient probabilistic 3{D} object detector for
  autonomous driving.
\newblock In {\em Proceedings of the IEEE Conference on Computer Vision and
  Pattern Recognition (CVPR)}, 2019.

\bibitem{mittal2020just}
Himangi Mittal, Brian Okorn, and David Held.
\newblock Just go with the flow: Self-supervised scene flow estimation.
\newblock In {\em Proceedings of the IEEE/CVF Conference on Computer Vision and
  Pattern Recognition}, pages 11177--11185, 2020.

\bibitem{REF:StarNet_2019}
Jiquan Ngiam, Benjamin Caine, Wei Han, Brandon Yang, Yuning Chai, Pei Sun, Yin
  Zhou, Xi Yi, Ouais Alsharif, Patrick Nguyen, Zhifeng Chen, Jonathon Shlens,
  and Vijay Vasudevan.
\newblock Starnet: Targeted computation for object detection in point clouds.
\newblock {\em CoRR}, 2019.

\bibitem{owoyemi2018spatiotemporal}
Joshua Owoyemi and Koichi Hashimoto.
\newblock Spatiotemporal learning of dynamic gestures from 3d point cloud data.
\newblock In {\em 2018 IEEE International Conference on Robotics and Automation
  (ICRA)}, pages 1--5. IEEE, 2018.

\bibitem{prantl2019tranquil}
Lukas Prantl, Nuttapong Chentanez, Stefan Jeschke, and Nils Thuerey.
\newblock Tranquil clouds: Neural networks for learning temporally coherent
  features in point clouds.
\newblock {\em arXiv preprint arXiv:1907.05279}, 2019.

\bibitem{qi2020imvotenet}
Charles~R Qi, Xinlei Chen, Or Litany, and Leonidas~J Guibas.
\newblock Imvotenet: Boosting 3d object detection in point clouds with image
  votes.
\newblock In {\em Proceedings of the IEEE/CVF Conference on Computer Vision and
  Pattern Recognition}, pages 4404--4413, 2020.

\bibitem{qi2019deep}
Charles~R Qi, Or Litany, Kaiming He, and Leonidas~J Guibas.
\newblock Deep hough voting for 3d object detection in point clouds.
\newblock In {\em Proceedings of the IEEE International Conference on Computer
  Vision}, pages 9277--9286, 2019.

\bibitem{qi2018frustum}
Charles~R Qi, Wei Liu, Chenxia Wu, Hao Su, and Leonidas~J Guibas.
\newblock Frustum pointnets for 3d object detection from rgb-d data.
\newblock In {\em CVPR}, 2018.

\bibitem{qi2017pointnet}
Charles~R Qi, Hao Su, Kaichun Mo, and Leonidas~J Guibas.
\newblock Pointnet: Deep learning on point sets for 3d classification and
  segmentation.
\newblock {\em CVPR}, 2017.

\bibitem{qi2017pointnetplusplus}
Charles~R Qi, Li Yi, Hao Su, and Leonidas~J Guibas.
\newblock Pointnet++: Deep hierarchical feature learning on point sets in a
  metric space.
\newblock {\em arXiv preprint arXiv:1706.02413}, 2017.

\bibitem{russakovsky2015imagenet}
Olga Russakovsky, Jia Deng, Hao Su, Jonathan Krause, Sanjeev Satheesh, Sean Ma,
  Zhiheng Huang, Andrej Karpathy, Aditya Khosla, Michael Bernstein, et~al.
\newblock Imagenet large scale visual recognition challenge.
\newblock {\em International journal of computer vision}, 115(3):211--252,
  2015.

\bibitem{shi2020pv}
Shaoshuai Shi, Chaoxu Guo, Li Jiang, Zhe Wang, Jianping Shi, Xiaogang Wang, and
  Hongsheng Li.
\newblock Pv-rcnn: Point-voxel feature set abstraction for 3d object detection.
\newblock In {\em Proceedings of the IEEE/CVF Conference on Computer Vision and
  Pattern Recognition}, pages 10529--10538, 2020.

\bibitem{shi2018pointrcnn}
Shaoshuai Shi, Xiaogang Wang, and Hongsheng Li.
\newblock Pointrcnn: 3d object proposal generation and detection from point
  cloud.
\newblock {\em arXiv preprint arXiv:1812.04244}, 2018.

\bibitem{REF:Point-GNN_CVPR2020}
Weijing Shi and Ragunathan~(Raj) Rajkumar.
\newblock Point-gnn: Graph neural network for 3d object detection in a point
  cloud.
\newblock In {\em The IEEE Conference on Computer Vision and Pattern
  Recognition (CVPR)}, June 2020.

\bibitem{simony2018complex}
Martin Simony, Stefan Milzy, Karl Amendey, and Horst-Michael Gross.
\newblock Complex-yolo: An euler-region-proposal for real-time 3d object
  detection on point clouds.
\newblock In {\em Proceedings of the European Conference on Computer Vision
  (ECCV) Workshops}, pages 0--0, 2018.

\bibitem{REF:MVX-Net_ICRA2019}
Vishwanath~A. Sindagi, Yin Zhou, and Oncel Tuzel.
\newblock Mvx-net: Multimodal voxelnet for 3d object detection.
\newblock In {\em International Conference on Robotics and Automation, {ICRA}
  2019, Montreal, QC, Canada, May 20-24, 2019}, pages 7276--7282. {IEEE}, 2019.

\bibitem{REF:wbf2019}
Roman Solovyev, Weimin Wang, and Tatiana Gabruseva.
\newblock Weighted boxes fusion: ensembling boxes for object detection models.
\newblock {\em arXiv preprint arXiv:1910.13302}, 2019.

\bibitem{song2016deep}
Shuran Song and Jianxiong Xiao.
\newblock Deep sliding shapes for amodal 3d object detection in rgb-d images.
\newblock In {\em CVPR}, 2016.

\bibitem{sun2020scalability}
Pei Sun, Henrik Kretzschmar, Xerxes Dotiwalla, Aurelien Chouard, Vijaysai
  Patnaik, Paul Tsui, James Guo, Yin Zhou, Yuning Chai, Benjamin Caine, et~al.
\newblock Scalability in perception for autonomous driving: Waymo open dataset.
\newblock In {\em Proceedings of the IEEE/CVF Conference on Computer Vision and
  Pattern Recognition}, pages 2446--2454, 2020.

\bibitem{REF:tian2019fcos}
Zhi Tian, Chunhua Shen, Hao Chen, and Tong He.
\newblock {FCOS}: Fully convolutional one-stage object detection.
\newblock In {\em Proc. Int. Conf. Computer Vision (ICCV)}, 2019.

\bibitem{REF:VotingforVoting_RSS2015}
Dominic~Zeng Wang and Ingmar Posner.
\newblock Voting for voting in online point cloud object detection.
\newblock In {\em Proceedings of Robotics: Science and Systems}, Rome, Italy,
  July 2015.

\bibitem{wang2019makes}
Weiyao Wang, Du Tran, and Matt Feiszli.
\newblock What makes training multi-modal networks hard?
\newblock {\em arXiv preprint arXiv:1905.12681}, 2019.

\bibitem{REF:PillarNet_ECCV2020}
Yue Wang, Alireza Fathi, Abhijit Kundu, David Ross, Caroline Pantofaru, Thomas
  Funkhouser, and Justin Solomon.
\newblock Pillar-based object detection for autonomous driving.
\newblock In {\em ECCV}, 2020.

\bibitem{weng2019baseline}
Xinshuo Weng and Kris Kitani.
\newblock A baseline for 3d multi-object tracking.
\newblock {\em arXiv preprint arXiv:1907.03961}, 2019.

\bibitem{weng20204d}
Xinshuo Weng, Jianren Wang, Sergey Levine, Kris Kitani, and Nicholas Rhinehart.
\newblock 4d forecasting: Sequential forecasting of 100,000 points, 2020.

\bibitem{weng2020joint}
Xinshuo Weng, Ye Yuan, and Kris Kitani.
\newblock Joint 3d tracking and forecasting with graph neural network and
  diversity sampling.
\newblock {\em arXiv preprint arXiv:2003.07847}, 2020.

\bibitem{xie2020self}
Qizhe Xie, Minh-Thang Luong, Eduard Hovy, and Quoc~V Le.
\newblock Self-training with noisy student improves imagenet classification.
\newblock In {\em Proceedings of the IEEE/CVF Conference on Computer Vision and
  Pattern Recognition}, pages 10687--10698, 2020.

\bibitem{REF:pointfusion_CVPR2018}
Danfei Xu, Dragomir Anguelov, and Ashesh Jain.
\newblock {PointFusion}: Deep sensor fusion for 3d bounding box estimation.
\newblock In {\em Proceedings of the IEEE Conference on Computer Vision and
  Pattern Recognition (CVPR)}, 2018.

\bibitem{yalniz2019billion}
I~Zeki Yalniz, Herv{\'e} J{\'e}gou, Kan Chen, Manohar Paluri, and Dhruv
  Mahajan.
\newblock Billion-scale semi-supervised learning for image classification.
\newblock {\em arXiv preprint arXiv:1905.00546}, 2019.

\bibitem{REF:second_2018}
Yan Yan, Yuxing Mao, and Bo Li.
\newblock Second: Sparsely embedded convolutional detection.
\newblock {\em Sensors}, 18(10):3337, 2018.

\bibitem{REF:yang2018pixor}
Bin Yang, Wenjie Luo, and Raquel Urtasun.
\newblock Pixor: Real-time 3d object detection from point clouds.
\newblock In {\em Proceedings of the IEEE Conference on Computer Vision and
  Pattern Recognition}, pages 7652--7660, 2018.

\bibitem{yang20203dssd}
Zetong Yang, Yanan Sun, Shu Liu, and Jiaya Jia.
\newblock 3dssd: Point-based 3d single stage object detector.
\newblock In {\em Proceedings of the IEEE/CVF Conference on Computer Vision and
  Pattern Recognition}, pages 11040--11048, 2020.

\bibitem{REF:yang2018ipod}
Zetong Yang, Yanan Sun, Shu Liu, Xiaoyong Shen, and Jiaya Jia.
\newblock Ipod: Intensive point-based object detector for point cloud, 2018.

\bibitem{REF:STD_ICCV2019}
Z. {Yang}, Y. {Sun}, S. {Liu}, X. {Shen}, and J. {Jia}.
\newblock Std: Sparse-to-dense 3d object detector for point cloud.
\newblock In {\em 2019 IEEE/CVF International Conference on Computer Vision
  (ICCV)}, pages 1951--1960, 2019.

\bibitem{REF:HVNet2020}
M. {Ye}, S. {Xu}, and T. {Cao}.
\newblock Hvnet: Hybrid voxel network for lidar based 3d object detection.
\newblock In {\em 2020 IEEE/CVF Conference on Computer Vision and Pattern
  Recognition (CVPR)}, pages 1628--1637, 2020.

\bibitem{REF:Spatiotemporal_transformer_CVPR2020}
J. {Yin}, J. {Shen}, C. {Guan}, D. {Zhou}, and R. {Yang}.
\newblock Lidar-based online 3d video object detection with graph-based message
  passing and spatiotemporal transformer attention.
\newblock In {\em 2020 IEEE/CVF Conference on Computer Vision and Pattern
  Recognition (CVPR)}, pages 11492--11501, 2020.

\bibitem{zakharov2020autolabeling}
Sergey Zakharov, Wadim Kehl, Arjun Bhargava, and Adrien Gaidon.
\newblock Autolabeling 3d objects with differentiable rendering of sdf shape
  priors.
\newblock In {\em Proceedings of the IEEE/CVF Conference on Computer Vision and
  Pattern Recognition}, pages 12224--12233, 2020.

\bibitem{zhou2020end}
Yin Zhou, Pei Sun, Yu Zhang, Dragomir Anguelov, Jiyang Gao, Tom Ouyang, James
  Guo, Jiquan Ngiam, and Vijay Vasudevan.
\newblock End-to-end multi-view fusion for 3d object detection in lidar point
  clouds.
\newblock In {\em Conference on Robot Learning}, pages 923--932, 2020.

\bibitem{zhou2018voxelnet}
Yin Zhou and Oncel Tuzel.
\newblock Voxelnet: End-to-end learning for point cloud based 3d object
  detection.
\newblock In {\em CVPR}, 2018.

\bibitem{zou2019confidence}
Yang Zou, Zhiding Yu, Xiaofeng Liu, BVK Kumar, and Jinsong Wang.
\newblock Confidence regularized self-training.
\newblock In {\em Proceedings of the IEEE International Conference on Computer
  Vision}, pages 5982--5991, 2019.

\end{thebibliography}
